\documentclass{article}

\usepackage{PRIMEarxiv}

\usepackage[utf8]{inputenc} 
\usepackage[T1]{fontenc}    
\usepackage{hyperref}       
\usepackage{url}            
\usepackage{booktabs}       
\usepackage{amsfonts}       
\usepackage{nicefrac}       
\usepackage{microtype}      
\usepackage{lipsum}
\usepackage{fancyhdr}       
\usepackage{graphicx}       
\graphicspath{{media/}}     
\usepackage{amsmath}
\usepackage{caption}     
\usepackage{subcaption}  
\usepackage{tabularx}    
\usepackage{colortbl}
\usepackage{multirow}
\usepackage{bm}
\usepackage[most]{tcolorbox}

\usepackage[colorinlistoftodos]{todonotes}

\usepackage{tikz}
\usetikzlibrary{shapes.geometric, arrows}
\usetikzlibrary{arrows} 

\tikzstyle{startstop} = [cylinder, shape border rotate=90, aspect=.1, draw, minimum height=.6cm, minimum width=.6cm, text width=1.35cm, align=center]
\tikzstyle{process} = [rectangle, draw, fill=gray!20, minimum width=1.2cm, minimum height=.6cm, text width=1.2cm, align=center]
\tikzstyle{decision} = [diamond, draw, minimum width=.6cm, minimum height=.6cm]
\tikzstyle{arrow} = [thick, ->, >=stealth]
\tikzstyle{line} = [thick, -, >=stealth]
\tikzstyle{box} = [rectangle, draw, minimum width=1.2cm, minimum height=.4cm, text width=1.5cm, align=center]

\usetikzlibrary{calc}
\usetikzlibrary{arrows.meta, positioning}
\usepackage{float}
\title{Neurosymbolic Artificial Intelligence for Robust Network Intrusion Detection: From Scratch to Transfer Learning}

\author{
Huynh T. T. Tran, Jacob Sander, Achraf Cohen, Brian Jalaian \\
\textit{University of West Florida, Pensacola, USA} \\
\texttt{\{ht68, jhs39\}@students.uwf.edu}, \texttt{\{acohen, bjalaian\}@uwf.edu} \\
\and
\textbf{Nathaniel D. Bastian} \\
\textit{United States Military Academy, West Point, New York, USA} \\
\texttt{nathaniel.bastian@westpoint.edu}
}

\begin{document}

\maketitle

\begin{abstract}
Network Intrusion Detection Systems (NIDS) play a vital role in protecting digital infrastructures against increasingly sophisticated cyber threats. In this paper, we extend \textbf{ODXU}, a Neurosymbolic AI (NSAI) framework that integrates deep embedded clustering for feature extraction, symbolic reasoning using XGBoost, and comprehensive uncertainty quantification (UQ) to enhance robustness, interpretability, and generalization in NIDS. The extended ODXU incorporates score-based methods (e.g., Confidence Scoring, Shannon Entropy) and metamodel-based techniques, including SHAP values and Information Gain, to assess the reliability of predictions. Experimental results on the CIC-IDS-2017 dataset show that ODXU outperforms traditional neural models across six evaluation metrics, including classification accuracy and false omission rate. While transfer learning has seen widespread adoption in fields such as computer vision and natural language processing, its potential in cybersecurity has not been thoroughly explored. To bridge this gap, we develop a transfer learning strategy that enables the reuse of a pre-trained ODXU model on a different dataset. Our ablation study on ACI-IoT-2023 demonstrates that the optimal transfer configuration involves reusing the pre-trained autoencoder, retraining the clustering module, and fine-tuning the XGBoost classifier, and outperforms traditional neural models when trained with as few as 16,000 samples (approximately 50\% of the training data). Additionally, results show that metamodel-based UQ methods consistently outperform score-based approaches on both datasets.
\end{abstract}

\textbf{Keywords:} Network Intrusion Detection Systems, Neurosymbolic AI, Transfer Learning, Uncertainty Quantification, Metamodel, Open Set Recognition

\section{Introduction}

As digital infrastructure expands in scale and complexity, safeguarding networked systems against malicious intrusions remains a central challenge in cybersecurity. Network Intrusion Detection Systems (NIDS) serve as the first line of defense by continuously monitoring network traffic to detect and classify abnormal or malicious activity \cite{al2021intelligent}. While deep learning and other neural-based models have shown success in learning complex traffic patterns, they often fall short in generalizing to novel attack types and provide limited interpretability. These limitations hinder their deployment in high-stakes environments where adaptability and transparency are crucial.

Neurosymbolic Artificial Intelligence (NSAI) has emerged as a promising solution to these challenges. NSAI combines the expressive feature-learning capability of neural networks with the logical reasoning strengths of symbolic AI \cite{sander2024uncertainty,jalaian2023neurosymbolic}. This hybrid approach not only enhances robustness and interpretability but also enables structured decision-making through rule-based models. In particular, decision trees, such as those implemented in XGBoost, can be interpreted as propositional logic expressions, making them well-suited for symbolic reasoning within NSAI architectures.

However, a major limitation in deploying NSAI-based systems across real-world intrusion detection scenarios is the dependence on large, labeled datasets tailored to specific attack types or domains. Building a new model from scratch for each dataset or subtask can be computationally expensive and operationally inefficient. To address this challenge, we draw inspiration from the broader machine learning community, where \textit{transfer learning} has become a widely adopted strategy. Transfer learning allows pre-trained models to be adapted to new tasks using limited labeled data, and has achieved notable success in domains such as computer vision \cite{shahoveisi2023application}, natural language processing \cite{amiriparian2022deepspectrumlite, moon2014multimodal}, and medical imaging \cite{kim2022transfer}. Despite its success in other domains, transfer learning remains underutilized in cybersecurity, particularly in adapting NIDS models to new datasets and attack scenarios.

In addition, given the high-stakes nature of cybersecurity applications, it is equally important to quantify the uncertainty of model predictions. Accurate uncertainty quantification (UQ) allows analysts to assess the confidence of the system, especially when facing unfamiliar or borderline inputs. To this end, we integrate both score-based methods (e.g., Confidence Scoring and Shannon Entropy) and metamodel-based approaches, which learn to predict the correctness of the model's outputs based on both the classifier model's input and augmented features. This integration not only improves interpretability but also provides a framework for more reliable decision-making in real-time intrusion detection.

In this work, we first extend the NSAI-based framework called \textbf{ODXU}, \textbf{O}pen Set Recognition with \textbf{D}eep Embedded Clustering for \textbf{X}GBoost and \textbf{U}ncertainty Quantification \cite{sander2024uncertainty}. ODXU combines neural feature learning from packet payloads with symbolic reasoning through XGBoost and introduces new principled mechanisms for uncertainty quantification. We further develop a transfer learning approach that enables ODXU to be adapted efficiently to other datasets or detection tasks, enhancing its scalability and robustness. Our contributions are as follows:
\begin{itemize}
    \item We further investigate the ODXU framework, which integrates neural feature learning and symbolic reasoning to improve performance and interpretability in NIDS.
    \item We propose a transfer learning framework to adapt the ODXU model across different cybersecurity datasets and tasks, enhancing scalability and deployment efficiency.
    \item We incorporate and evaluate both score-based and metamodel-based UQ methods, improving the transparency and reliability of the intrusion detection system.
\end{itemize}

The remainder of this paper is organized as follows. Section~\ref{sec:background} provides the background information and related works. Section~\ref{sec:methodology} details the ODXU architecture and transfer learning framework. Section~\ref{sec:exp_setup} describes the experimental setup. Section~\ref{sec:results} presents and discusses our results. Section~\ref{sec:conclusion} concludes the paper and outlines future directions.

\section{Background and Related Works}
\label{sec:background}

This section provides an overview of key components underpinning our approach, including foundational work on NIDS, the potential of NSAI, the role of transfer learning in adapting models across domains, and the importance of UQ for robust and interpretable cybersecurity applications. Together, these areas form the basis for our ODXU architecture and transfer learning framework introduced in Section~\ref{sec:methodology}.

\subsection{Network Intrusion Detection Systems}

NIDS are essential components of modern cybersecurity frameworks. Their primary function is to inspect network traffic to identify suspicious or malicious behavior that may signal an ongoing or imminent cyberattack \cite{al2021intelligent}. Early NIDS implementations, such as those introduced by Denning in 1987 \cite{denning1987intrusion}, relied on signature-based detection, wherein observed traffic patterns were matched against known attack signatures. Although effective against previously identified threats, these systems are inherently limited when faced with novel or evolving attacks due to their static rule sets and dependence on extensive labeled data.

To overcome these limitations, machine learning (ML) and deep learning (DL) models have been widely adopted for data-driven anomaly detection \cite{al2021intelligent, sander2024uncertainty}. These models learn patterns of normal traffic and detect deviations that may correspond to malicious activity, improving generalization to previously unseen threats. In particular, the Open Set Recognition (OSR) framework has gained traction in the NIDS domain \cite{Yang2024}, as it enables classification of ``known known'' classes (benign and known attacks) while flagging ``unknown unknown'' instances (potential novel attacks).

Network traffic can be analyzed using either flow-level metadata or raw packet-level data. Recent studies have increasingly focused on packet-level payload data for its fine-grained temporal resolution and real-time detection potential \cite{farrukh2022payloadbyte}. Prominent datasets supporting research in this space include University of New South Wales Network-Based 2015 (UNSW-NB15) \cite{moustafa2015unsw}, Canadian Institute for Cybersecurity Intrusion Detection System 2017 (CIC-IDS-2017) \cite{sharafaldincicids2017}, and Army Cyber Institute Internet of Things Network Traffic Dataset 2023 (ACI-IOT-2023) \cite{Nack_2024b}.

\subsection{Neurosymbolic AI}

Neurosymbolic AI (NSAI) combines the high-capacity pattern recognition abilities of neural networks with the transparent, rule-based reasoning of symbolic AI. This hybridization is particularly attractive for domains such as cybersecurity, where models must handle vast, high-dimensional data while producing interpretable outputs. NSAI addresses key limitations of black-box ML/DL models, most notably, the lack of interpretability, by embedding symbolic structures like rules or trees within the learning process \cite{jalaian2023neurosymbolic, sander2024uncertainty}.

In the context of NIDS, NSAI enables both accurate classification and human-readable explanations, providing cybersecurity analysts with actionable insights \cite{bizzarri2024synergistic}. A common symbolic implementation in NSAI is the use of tree-based models, ranging from individual Decision Trees to boosted ensembles such as XGBoost. These models operate through IF-THEN rules derived from decision paths, offering a natural integration point for propositional logic. When combined with neural architectures that learn discriminative features from network payloads, NSAI enables structured, interpretable, and adaptive intrusion detection.

\subsection{Transfer Learning Techniques}

Transfer learning is a well-established approach in ML that aims to improve model performance on a target task by leveraging knowledge from related source tasks. This technique is especially beneficial when the target domain has limited labeled data or exhibits domain shift. In fields such as computer vision, natural language processing, and medical imaging, transfer learning has demonstrated considerable success \cite{shahoveisi2023application, amiriparian2022deepspectrumlite, moon2014multimodal, kim2022transfer, lee2022surgical, liu2021transtailor, guo2019spottune}.

Various strategies have been proposed, including:
\begin{itemize}
    \item \textit{Surgical fine-tuning: }adjusting selected layers of a neural network to align with the target distribution \cite{lee2022surgical}.
    \item \textit{Architecture pruning:} as in TransTailor, which simplifies the model to better fit task-specific constraints \cite{liu2021transtailor}.
    \item \textit{Adaptive fine-tuning: }such as SpotTune, which dynamically determines which layers to fine-tune per instance \cite{guo2019spottune}.
\end{itemize}

Despite its demonstrated efficacy, transfer learning remains underexplored in the cybersecurity domain, particularly for tasks involving NIDS. Current NIDS models are typically retrained from scratch for each new dataset or attack type, resulting in high resource costs and limited scalability. Integrating transfer learning with NSAI offers a promising path forward, allowing for more adaptable, efficient, and generalizable intrusion detection frameworks.

\subsection{Uncertainty Quantification}

Uncertainty Quantification (UQ) plays a critical role in trustworthy AI systems, especially in high-stakes domains like cybersecurity. UQ techniques assess the confidence of model predictions and are useful for flagging potentially erroneous outputs, detecting out-of-distribution samples, or identifying novel (unknown) attack types \cite{vadera2020ursabench}. In the NIDS context, UQ enhances situational awareness by helping analysts differentiate between confident and uncertain classifications. UQ approaches can be categorized into:
\begin{itemize}
    \item \textit{Score-based methods:} such as Confidence Scoring and Shannon Entropy \cite{hendrycks2016misclassificationood, wong2023uncertaintywintersim}, which assess uncertainty directly from classifier's output probabilities.
    \item \textit{Metamodel-based methods:} which involves training a secondary model to predict whether the classifier's prediction is likely to be correct \cite{ghosh2021ibmuncertainty, chen2019ibmwhitebox}.
\end{itemize}

In metamodel-based methods, our work adopts post-hoc, deterministic UQ techniques, which avoid retraining the classifier model and offer efficiency. The classifier model’s input is augmented either with output probabilities \cite{chen2019ibmwhitebox} or with interpretability-derived features, such as Shapley Additive Explanations (SHAP) values \cite{lundberg2017unified} and Information Gain (IG) \cite{quinlan1986induction}, to train a secondary model that estimates prediction confidence \cite{ghosh2021ibmuncertainty}.

\section{Methodology}
\label{sec:methodology}

This section outlines the method for processing network data, the design of the ODXU architecture, including the processing of raw packet-level data and the integration of Deep Embedded Clustering and XGBoost within the ODXU. We then provide the transfer learning strategies applied to ODXU, and the UQ mechanisms.

\subsection{Network Data Processing}

We use packet-level data in PCAP (Packet CAPture) format, which contains complete raw payloads and metadata. Compared to flow-level data that only summarizes aggregated traffic features (e.g., byte counts, duration), PCAP provides fine-grained, content-level information, allowing for more precise detection of malicious behavior in network traffic \cite{farrukh2022payloadbyte, bizzarri2024synergistic}. However, PCAP data is unstructured and unlabeled in most NIDS datasets.

To address this, we use Payload-Byte \cite{farrukh2022payloadbyte}, an open-source tool that converts each packet's payload into a standardized 1500-byte feature vector, independent of protocol. Headers are discarded to focus solely on application-level content. This process enables compatibility across datasets and facilitates learning from raw data. Payload-Byte also automatically labels packets, which supports supervised training of downstream models.

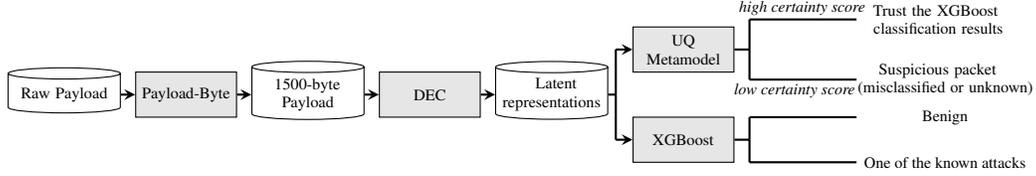
\begin{figure}[t!]
\centering
\tiny
\begin{tikzpicture}[node distance=0cm]

    \node (start) [startstop] {Raw Payload};
    \node (byte) [process, right of=start, xshift=1.62cm] {Payload-Byte};
    \node (payload) [startstop, right of=byte, xshift=1.62cm] {1500-byte Payload};
    \node (dec) [process, right of=payload, xshift=1.62cm] {DEC};
    \node (latent) [startstop, right of=dec, xshift=1.62cm] {\hspace{.05cm} Latent \hspace{.3cm} representations};

    \node (inter) [coordinate] at ($(latent.east)+(.1cm,0)$) {};

    \node (line1) [coordinate] at ($(inter.north)+(0,.6cm)$) {};
    \node (uq) [process, right of=line1, xshift=.9cm] {UQ \hspace{2cm} Metamodel};
    
    \node (inter1) [coordinate] at ($(uq.east)+(.2cm,0)$) {};
    \node (line2) [coordinate] at ($(inter1.north)+(0,.4cm)$) {};
    \node (line3) [right of=line2, xshift=1.5cm] {};
    \node (text1) [above of=line3, xshift = -.8cm, yshift = .15cm] {\textit{high certainty score}};
    \node (high) [right of=line3, xshift=1.15cm, text width=2cm] {Trust the XGBoost classification results};

    \node (line4) [coordinate] at ($(inter1.north)+(0,-.4cm)$) {};
    \node (line5) [right of=line4, xshift=1.5cm] {};
    \node (text2) [below of=line5, xshift = -.9cm, yshift = -.15cm] {\textit{low certainty score}};
    \node (low) [right of=line5, xshift=1.15cm, text width=2.4cm] {\hspace{.2cm} Suspicious packet \hspace{.2cm} (misclassified or unknown)};

    \node (line6) [coordinate] at ($(inter.north)+(0,-.6cm)$) {};
    \node (line7) [right of=line6, xshift=.6cm] {};
    \node (xgb) [process, right of=line7, xshift=.3cm] {XGBoost};

    \node (inter2) [coordinate] at ($(xgb.east)+(.2cm,0)$) {};
    \node (line8) [coordinate] at ($(inter2.north)+(0,.3cm)$) {};
    \node (line9) [right of=line8, xshift=1.5cm] {};  
    \node (benign) [right of=line9,  xshift=1.1cm] {Benign};

    \node (line10) [coordinate] at ($(inter2.south)+(0,-.3cm)$) {};
    \node (line11) [right of=line10, xshift=1.5cm] {};  
    \node (attack) [right of=line11,  xshift=1.1cm] {One of the known attacks};

    \draw [arrow] (start) -- (byte);
    \draw [arrow] (byte) -- (payload);
    \draw [arrow] (payload) -- (dec);
    \draw [arrow] (dec) -- (latent);

    \draw [line]  (latent) -- (inter);
    
    \draw [line] (inter) -- (line1);
    \draw [arrow] (line1) -- (uq);

    \draw [line] (uq) -- (inter1);
    \draw [line] (inter1) -- (line2);
    \draw [line] (line2) -- (line3);

    \draw [line] (inter1) -- (line4);
    \draw [line] (line4) -- (line5);
    
    \draw [line] (inter) -- (line6);
    \draw [arrow] (line6) -- (xgb);

    \draw [line] (xgb) -- (inter2);
    \draw [line] (inter2) -- (line8);
    \draw [line] (line8) -- (line9);

    \draw [line] (inter2) -- (line10);
    \draw [line] (line10) -- (line11);

\end{tikzpicture}
\vspace{-0.5em}
\caption{The architecture of the ODXU model}
\label{fig:pipeline}
\end{figure}

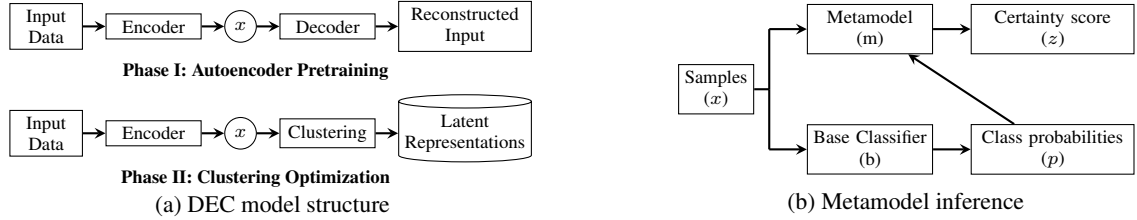
\begin{figure}[t!]
    \centering
    \begin{subfigure}[b]{.52\textwidth}
    \begin{minipage}{\textwidth}
        \centering
        \scriptsize
        \begin{tikzpicture}[node distance=0cm, every node/.style={align=center}]

        \node (input1) [draw, rectangle, text width = .8cm] {Input \\ Data};
        \node (encoder1) [draw, rectangle, right=.3cm of input1, text width = 1.1cm] {Encoder};
        \node (latent1) [draw, circle, right=.3cm of encoder1] {$x$};
        \node (decoder) [draw, rectangle, right=.3 cm of latent1, text width = 1.1cm] {Decoder};
        \node (recon) [draw, rectangle, right=.3 cm of decoder, , text width = 1.6cm] {Reconstructed \\ Input};

        \draw[arrow] (input1) -- (encoder1);
        \draw[arrow] (encoder1) -- (latent1);
        \draw[arrow] (latent1) -- (decoder);
        \draw[arrow] (decoder) -- (recon);

        \node at ($(input1)!.5!(recon)+(0,-.6cm)$) {\textbf{Phase I: Autoencoder Pretraining}};

        \node (input2) [draw, rectangle, below=.8cm of input1, text width = .8cm] {Input \\ Data};
        \node (encoder2) [draw, rectangle, right=.3cm of input2, text width = 1.1cm] {Encoder};
        \node (latent2) [draw, circle, right=.3cm of encoder2] {$x$};
        \node (cluster) [draw, rectangle, right=.3cm of latent2, text width = 1.1cm] {Clustering};
        \node (latent) [startstop, right=.3cm of cluster, , text width = 1.6cm] {Latent \\ Representations};

        \draw[arrow] (input2) -- (encoder2);
        \draw[arrow] (encoder2) -- (latent2);
        \draw[arrow] (latent2) -- (cluster);
        \draw[arrow] (cluster) -- (latent);

        \node at ($(input2)!.5!(latent)+(0,-.6cm)$) {\textbf{Phase II: Clustering Optimization}};

    \end{tikzpicture}
    \vspace{-.2cm}
        \caption{DEC model structure}
        \label{fig:dec_model}
    \end{minipage}
    \end{subfigure}
    \hfill
    \begin{subfigure}[b]{.46\linewidth}
    \begin{minipage}{\textwidth}
        \centering
        \scriptsize
        \begin{tikzpicture}[node distance=0, every node/.style={align=center}]

            \node (samples) [draw, rectangle] {Samples \\ ($x$)};
            \node (inter) [coordinate] at ($(samples.east)+(.2cm,0)$) {};
            \node (line1) [coordinate] at ($(inter.north)+(0cm,.8cm)$) {};
            \node (metamodel) [draw, rectangle, right=.5cm of line1, text width = 1.5cm] {Metamodel \\ (m)};
            \node (zscore) [draw, rectangle, right=.5cm of metamodel, text width = 2cm] {Certainty score \\ ($z$)};

            \node (line2) [coordinate] at ($(inter.south)+(0,-.8cm)$) {};
            \node (basemodel) [draw, rectangle, right=.5cm of line2, text width = 1.5cm] {Base Classifier \\ (b)};
            \node (probability) [draw, rectangle, right=.5cm of basemodel, text width = 2cm] {Class probabilities \\ ($p$)};

            \draw [line] (samples) -- (inter);
            \draw [line] (inter) -- (line1);
            \draw [arrow] (line1) -- (metamodel);
            \draw [arrow] (metamodel) -- (zscore);

            \draw [line] (inter) -- (line2);
            \draw [arrow] (line2) -- (basemodel);
            \draw [arrow] (basemodel) -- (probability);

            \draw [arrow] (probability) -- (metamodel);
            
        \end{tikzpicture}
        \caption{Metamodel inference}
        \label{fig:metamodel}
    \end{minipage}
    \end{subfigure}
    \caption{General structure of (a) DEC and (b) metamodel}
    \label{fig:general_structure}
\end{figure}

\subsection{Architecture of ODXU}

Figure~\ref{fig:pipeline} illustrates the overall architecture of ODXU. It consists of two core components: Deep Embedded Clustering and XGBoost.

\textit{Deep Embedded Clustering (DEC)} is a neural network-based clustering technique that jointly learns low-dimensional feature representations and cluster assignments \cite{xie2016dec}. The model consists of two phases, as shown in Figure~\ref{fig:dec_model}:

\textbf{Phase I -- Autoencoder  (AE):} A stacked denoising autoencoder is trained to reconstruct the original 1500-byte payload input. The encoder compresses the input into a lower-dimensional latent representation, while the decoder reconstructs the input from this latent space. This phase enables the encoder to learn meaningful structure in the data.

\textbf{Phase II -- Clustering Optimization:} The decoder is discarded, and a clustering layer is attached to the encoder output. Using the Student's t-distribution as a kernel, the model computes soft assignments between embedded points and centroids. An auxiliary target distribution is then defined to emphasize high-confidence assignments and refine the encoder and cluster centers by minimizing Kullback-Leibler (KL) divergence (Eq. \eqref{eq:KL}). 

In our model, we add contrastive loss as the inverse distance between centroids (Eq. \eqref{eq:contrastive}), and cross-entropy loss of classes (Eq. \eqref{eq:CE}), to the KL Divergence to enhance the accuracy of DEC in learning latent representations. Thus, the DEC objective function incorporates three loss terms as follows:
\begin{equation}
\mathcal{L} = L_{\text{KL}} + L_{\text{contrastive}} + L_{\text{CE}} \label{eq:DEC_loss},
\end{equation}
where
\begin{subequations}
\begin{align}
L_{\text{KL}} &= KL(P \parallel Q) \label{eq:KL} = \sum_{x} P(x) \log\left(\frac{P(x)}{Q(x)}\right),
\\
L_{\text{contrastive}} &= \frac{n_c(n_c-1)}{\sum_{i,j} \| u_i - u_j \|_2}, \label{eq:contrastive} \\
L_{\text{CE}} &= -\frac{1}{N} \sum_{i=1}^{N} y_i \log(\hat{y_i}), \label{eq:CE}
\end{align}
\end{subequations}
where $P$ and $Q$ denote the auxiliary and predicted soft cluster assignments, respectively; $n_c$ is the number of centroids; $u_i$ are cluster centroids; and $y_i$, $\hat{y}_i$ are ground truth and predicted class labels of sample $i$-th.

\textit{XGBoost} is the symbolic reasoning component in ODXU \cite{chen2016xgboost}, a high-performance gradient boosting tree algorithm. XGBoost operates on the latent representation outputs from DEC and performs multiclass classification to assign each packet to a specific known attack or benign class. Its tree-based logic provides interpretable IF-THEN rules, making it suitable for integration into NSAI frameworks.

\subsubsection{Evaluation Metrics}

To evaluate the performance of attack classification tasks in NIDS, we use the following metrics: 

\begin{itemize}
    \item Multiclass Classification Accuracy: The number of correctly classified samples divided by the total number of samples.
    \item Binary Classification Accuracy: The accuracy of the model when determining if the samples are of ``any type of known attack'' or ``benign.''
    \item  Misclassified Positive Rate: The proportion of attacks correctly flagged as an attack but misclassified as to which specific attack type.
    \item False Omission Rate: The proportion of attacks incorrectly classified as benign.
    \item F1 Score: The F1 Score in the binary case - any type of known attack, or benign.
    \item Competence: The sum of the certainty scores of the true positives minus the false positives, divided by a number of total positives, in the binary case.
\end{itemize}
These are referenced from \cite{wong2023uncertaintywintersim}, and provide a comprehensive assessment of our system's classification performance on NIDS datasets.

\begin{table}[ht!]
    \caption{Transfer Learning Scenarios; FT: fine-tune}
    \label{tab:scenarios}
    \centering
    \renewcommand{\arraystretch}{1.15}
    \begin{tabularx}{.75\linewidth}{>{\hsize=1\hsize}X*{6}{>{\centering\arraybackslash}p{.8cm}}}
    \hline
    \multirow{2}{*}{\textbf{ODXU Components}} & \multicolumn{6}{c}{\textbf{Case}} \\
    \cline{2-7}
    & \textbf{1} & \textbf{2} & \textbf{3} & \textbf{4} & \textbf{5} & \textbf{6} \\
    \hline
    AE                   & FT     & As is  & As is  & FT     & As is  & As is  \\
    Clustering           & Train  & FT     & Train  & Train  & FT     & Train  \\
    Classifier (XGBoost) & Train  & Train  & Train  & FT     & FT     & FT     \\
    \hline
    \end{tabularx}
\end{table}

\subsection{Transfer Learning of ODXU}

To enable ODXU to generalize across different datasets or intrusion detection tasks, we propose a transfer learning framework that explores the reuse and adaptation of its components. We pose two research questions:
\begin{enumerate}
    \item Which components of the ODXU architecture (e.g., AE, clustering, XGBoost) should be fine-tuned or trained for effective transfer learning?
    \item How many labeled samples are required to outperform baseline machine learning models such as Fully Connected Neural Networks (FcNN) or 1D Convolutional Neural Networks (1D-CNN)?
\end{enumerate}

To answer these questions, we designed an experiment using the six transfer learning scenarios as shown in Table~\ref{tab:scenarios}.
The AE has two options: ``As is,'' where the pre-trained AE from a source dataset is loaded and used without any further training, and ``FT'' (fine-tune), where the pre-trained AE is loaded and then further trained on a target dataset. The clustering component also considers two options: In the ``FT'' option, it loads a pre-trained clustering model from the source dataset and fine-tunes it on the target dataset. In the ``Train'' option, it loads the parameters of the pre-trained AE instead of the pre-trained clustering model and trains these parameters on the target dataset\footnote{Note: As depicted in Figure~\ref{fig:dec_model}, the clustering module is initialized either from its pre-trained parameters of source dataset or from the encoder of the AE. Therefore, if the AE is fine-tuned, the clustering cannot be initialized from its pre-trained checkpoint, as the encoder has changed. As a result, combinations such as FT-FT-Train or FT-FT-FT are invalid and excluded from our experiments.}.
The classifier has two options: ``FT,'' where a pre-trained classifier is loaded and fine-tuned on the target dataset, and ``Train,'' where the classifier is trained entirely from scratch.

\subsection{Uncertainty Quantification Methods}

We employ five different uncertainty quantification (UQ) methods in our analysis, including two scoring-based methods, Confidence Scoring and Shannon Entropy \cite{smith2011quantifying}, as well as three variants of metamodel-based approaches.

The metamodel-based UQ framework builds upon a base classifier $b(.)$ and a secondary binary classifier $m(.)$, which is trained to predict the reliability of the base model’s outputs. The metamodel input, denoted $\boldsymbol{X_{\text{MetaUQ}}}$, comprises the original feature set $\boldsymbol{X_b}$ along with augmented features derived from the base model. The target label for training the metamodel, $\boldsymbol{y_m}$, indicates whether the base classifier's prediction is correct (label 0) or incorrect (label 1), and is defined as:
\begin{align}
\label{target}
\boldsymbol{y_m} = 
\begin{cases} 
0 & \text{if }  b(x_b) = y_b\\
1 & \text{if } b(x_b) \neq y_b,
\end{cases}
\end{align}
where $y_b$ is the ground truth label and $b(x_b)$ is the predicted output for sample $x_b$.

The metamodel $m(.)$ learns to estimate $z = m(x_m)$ by minimizing the discrepancy between its output and the true correctness label $\boldsymbol{y_m}$. The resulting value $z$ can be interpreted as the probability that the base classifier’s prediction is correct, thereby serving as an estimate of prediction confidence. A general interface of the metamodel-based approach is shown in Figure~\ref{fig:metamodel}.

\subsubsection{Confidence Scoring}

Confidence scoring offers a lightweight and direct method for approximating prediction certainty. This score is derived by taking the difference between the top two predicted class probabilities:
\begin{equation}
    z_\text{conf}(x) = p_{(k)} - p_{(k-1)},
    \label{eqn:confidence}
\end{equation}
where $z_\text{conf}(x)$ represents the confidence score for input $x$, and $p_{(k)}$ and $p_{(k-1)}$ are the highest and second highest probabilities in the set of class probabilities $\boldsymbol{p}$, respectively. A larger difference reflects higher model confidence, whereas a smaller value indicates increased uncertainty.

\subsubsection{Shannon Entropy}

Shannon entropy is a fundamental concept in information theory, which is often utilized to assess uncertainty. This method computes the entropy for each sample based on the predicted probabilities across all classes produced by the base classifier. The entropy of a specific sample $x$ is determined using the formula below:
\begin{equation}
    z_\text{entropy}(x) = -\sum p_i \log(p_i),
    \label{eqn:entropy}
\end{equation}
where $p_i$ is the predicted probability of class $i$, and the summation spans all possible classes. Unlike confidence score, higher entropy values denote greater uncertainty, while lower values suggest more confident predictions.

\subsubsection{MetaUQ}

The MetaUQ approach utilizes a metamodel that is augmented with additional information beyond the original features, enabling it to infer uncertainty from patterns not explicitly available to the base classifier. In this paper, we considered three metamodels:

\paragraph{$\text{MetaUQ}_{prob}$} In this metamodel, we augmented the base classifier input with both the sorted class probabilities $\boldsymbol{p'}$ and the confidence score from Eq.~\eqref{eqn:confidence}. While the confidence score captures the gap between the top two class probabilities, the full sorted vector reflects the distribution across all classes, potentially revealing ambiguous or flat distributions that indicate uncertainty \cite{tagasovska2019single}. The augmented input to the $\text{MetaUQ}_{prob}$ is thus:
\begin{equation}
    \boldsymbol{X}_{\text{MetaUQ}_{prob}} = [\boldsymbol{X_b}, \boldsymbol{p'}, \boldsymbol{z_\text{conf}}].
\end{equation}

\paragraph{$\text{MetaUQ}_{SHAP}$} Shapley Additive Explanations (SHAP) \cite{lundberg2017unified} were originally introduced in cooperative game theory to assign value to individual contributors within a coalition. In recent years, SHAP has gained widespread use in machine learning as a tool for interpreting model predictions \cite{lundberg_shap}. In this approach, each feature in the input vector $x$ is viewed as a player in a coalition, and the goal is to quantify its individual contribution to the model's output.

For decision tree-based models such as XGBoost, SHAP values can be computed in polynomial time using algorithms that exploit the tree structure \cite{lundberg2020local}. This enables the generation of local explanations, i.e., the impact of each feature on a specific prediction, without relying on sampling or approximations. In our implementation, we compute SHAP values for each input sample $\boldsymbol{x}$ as follows \cite{lundberg2017unified}:
\begin{equation}
    \phi_i(b, x) = \sum_{S \subseteq I \setminus \{i\}} \frac{|S|! (|I|-|S|-1)!}{|I|!} \left[b(S \cup \{i\}) - b(S)\right],
    \label{eqn:shapvalue}
\end{equation}
where $\phi_i(b,x)$ is the SHAP value for feature $i$ with respect to the base classifier $b(.)$ and input $\boldsymbol{x}$. The operator $!$ denotes factorial, and $|\cdot|$ indicates the size of a set. The set $I$ contains all input features, while $S \subseteq I \setminus \{i\}$ represents a subset of features excluding $i$. The term $b(S \cup \{i\}) - b(S)$ reflects the marginal effect of including feature $i$ in subset $S$.

Following class-specific interpretation practices \cite{lundberg_shap}, we then extract SHAP values corresponding to the predicted class. The final augmented input to the $\text{MetaUQ}_{SHAP}$ is thus:
\begin{equation}
    \boldsymbol{X}_{\text{MetaUQ}_{SHAP}} = [\boldsymbol{X_b}, \boldsymbol{\phi}(b,x)].
    \label{eqn:shap_augment}
\end{equation}

\paragraph{$\text{MetaUQ}_{IG}$} Information Gain (IG) quantifies how much predictive uncertainty is reduced when splitting on a given feature. In tree models like XGBoost, the gain from a split is calculated as \cite{chen2016xgboost}:
\begin{equation}
\mathcal{L}_{\text{split}} = \frac{1}{2} \left[ 
\frac{\left( \sum_{i \in I_L} g_i \right)^2}{\sum_{i \in I_L} h_i + \lambda} 
+ 
\frac{\left( \sum_{i \in I_R} g_i \right)^2}{\sum_{i \in I_R} h_i + \lambda} 
- 
\frac{\left( \sum_{i \in I} g_i \right)^2}{\sum_{i \in I} h_i + \lambda} 
\right] - \gamma,
\label{eqn:IG_XGBoost}
\end{equation}
where $I_L$ and $I_R$ are the left-hand side and right-hand side child nodes, $g_i$ and $h_i$ are the Gradient and Hessian of the loss for sample $i$, and $\lambda$ and $\gamma$ are regularization hyperparameters. The cumulative gain for each feature across all splits can be computed as:
\begin{equation}
    \text{Gain}(f) = \sum_{\text{\# splits using } f} \mathcal{L}_\text{split}.
\end{equation}
These scores are replicated to all features and samples, forming a matrix $\text{IG}_{\text{matrix}}$. We also append the sorted class probabilities $\boldsymbol{p'}$ to form the final augmented input:
\begin{equation}
    \boldsymbol{X}_{\text{MetaUQ}_{IG}} = [\boldsymbol{X_b},\boldsymbol{p'}, \text{IG}_{\text{matrix}}].
    \label{eqn:IG_augment}
\end{equation}

\subsubsection{UQ Evaluation Metrics}

We assess the performance of UQ methods on two primary tasks: detecting misclassifications and OSR. For the misclassification task, we assigned each test sample a binary label: 0 if correctly classified, and 1 if misclassified. We then use an Area Under the Receiver Operating Characteristic (AUROC) curve to measure how well uncertainty scores can distinguish between the two \cite{bradley1997auroc}. 

For the OSR task, we exclude certain attack types from training and treat them as ``unknown'' during inference. We label these unknowns as 1 and known types as 0, and again evaluate AUROC using the uncertainty scores \cite{vadera2020ursabench}.

In addition to AUROC, we report True Positive at a fixed True Negative rate of .95 (TP@TN=.95), which quantifies detection performance under a rigorous false positive constraint, reflecting operational needs in cybersecurity settings \cite{matejek2024safeguarding}.

\section{Experimental Setup}
\label{sec:exp_setup}

\subsection{Datasets}
\label{subsec:datasets}

\paragraph{CIC-IDS-2017.} The Canadian Institute for Cybersecurity Intrusion Detection Systems 2017 dataset \cite{sharafaldincicids2017} is a large-scale benchmark that closely resembles real-world network traffic scenarios. It comprises approximately 8 million labeled samples, encompassing both benign and a diverse range of contemporary cyberattacks, including DDoS, DoS Hulk, Port Scan, Brute Force, and others (a total of 15 classes). The descriptive statistics of CIC-IDS-2017 are presented in Table~\ref{tab:cic_ids_2017_class_stats}.

\begin{table}[ht!]
\centering
\caption{Descriptive statistics of class distribution in the CIC-IDS-2017 dataset.}
\label{tab:cic_ids_2017_class_stats}
\renewcommand{\arraystretch}{1.15}
\begin{tabular}{|l|r|r|}
\hline
\textbf{Class} & \textbf{Number of Samples} & \textbf{Percent (\%)} \\
\hline
Benign & 3,328,591 & 43.52 \\
DoS Hulk & 2,219,061 & 29.03 \\
DDoS & 618,544 & 8.09 \\
SSH-Patator & 181,147 & 2.37 \\
FTP-Patator & 110,636 & 1.45 \\
Infiltration & 41,725 & .55 \\
Heartbleed & 41,283 & .54 \\
DoS GoldenEye & 34,293 & .45 \\
Web Attack – Brute Force & 28,920 & .38 \\
DoS slowloris & 20,877 & .27 \\
DoS Slowhttptest & 9,778 & .13 \\
Web Attack – XSS & 6,767 & .09 \\
Bot & 5,143 & .07 \\
PortScan & 946 & .01 \\
Web Attack – Sql Injection & 45 & .00 \\
\hline
\textbf{Total} & \textbf{7,647,755} & \textbf{100.00} \\
\hline
\end{tabular}
\end{table}

\paragraph{ACI-IoT-2023.} The Army Cyber Institute Internet of Things Network Traffic Dataset 2023 \cite{aciiot2023} provides recent and comprehensive coverage of intrusion scenarios in IoT environments, making it a strong benchmark for evaluating model generalization in cross-domain settings. It includes ten classes, featuring a dominant share of benign alongside several rare but critical attack types such as DNS Flood, Vulnerability Scan, and OS Scan. The class distribution of ACI-IoT-2023 is shown in Table~\ref{tab:aci_iot2023_class_stats}.

\begin{table}[ht!]
\centering
\caption{Descriptive statistics of class distribution in the ACI-IoT-2023 dataset.}
\label{tab:aci_iot2023_class_stats}
\renewcommand{\arraystretch}{1.15}
\begin{tabular}{|l|r|r|}
\hline
\textbf{Class} & \textbf{Samples} & \textbf{Percent (\%)} \\
\hline
Benign              & 601,868 & 95.31 \\
DNS Flood           & 18,577  & 2.94 \\
Dictionary Attack   & 4,645   & .74 \\
Slowloris           & 2,974   & .47 \\
SYN Flood           & 2,113   & .33 \\
Port Scan           & 582       & .09 \\
Vulnerability Scan  & 445       & .07 \\
OS Scan             & 156       & .02 \\
UDP Flood           & 68        & .01 \\
ICMP Flood          & 58        & .01 \\
\hline
\textbf{Total}      & \textbf{631,486} & \textbf{100.00} \\
\hline
\end{tabular}
\end{table}

\subsection{Computational Environment}

All experiments were conducted on a system equipped with dual Intel® Xeon® Gold 5218R CPUs (2.10 GHz), providing 80 logical threads and 754 GiB of system memory. Model training and inference were accelerated using a single NVIDIA A40 GPU (Ampere architecture, 48 GB memory) out of eight available. Runtime was monitored to evaluate computational efficiency and scalability.

\subsection{Attack Recognition}
\label{subsec:attack_recognition}

To prepare the CIC-IDS-2017 dataset, raw PCAP files were parsed using the Payload-Byte tool to extract 1500-byte payload representations per packet. To mitigate class imbalance, Benign and DoS Hulk classes were downsampled by 90\%, and DDoS by 67\%. The resulting dataset was split 50/50 into \textit{DEC-Train} and \textit{DEC-Test}. A DEC model was trained on \textit{DEC-Train} (split into 75/25 for training and validation) to infer a 12-dimensional latent representation, which was then applied to \textit{DEC-Test}. This reduced set was further split equally into \textit{XGBoost-Train} and \textit{XGBoost-Test} to train the XGBoost classifier and ensure no data leakage. For baseline comparisons, FcNN and 1D-CNN models were trained on the original Payload-Byte format using the combined \textit{DEC-Train} and \textit{XGBoost-Train} sets. The FcNN consisted of three fully connected layers of sizes [1024, 512, 68], totaling 2,097,743 parameters. The 1D-CNN had convolutional layers with channels [32, 64, 128] (kernel size 3) and a dense layer of 75 neurons, totaling 185,759 parameters.

For the ACI-IoT-2023 dataset, the benign class was downsampled by 95\%, while rare attack classes like ICMP Flood and UDP Flood were upsampled by 200\%. The dataset was then split 50/50 into \textit{DEC-Train} and \textit{DEC-Test}. We used four training configurations (10\%, 25\%, 50\%, and 75\% of \textit{DEC-Train}) to analyze data efficiency. Each subset was also split into 75/25 for training and validation. The \textit{DEC-Test} set, transformed into 12 latent features by DEC, was split equally into \textit{XGBoost-Train} and \textit{XGBoost-Test} for classifier training. Here, FcNN used [1024, 512, 100] layers with 2,115,180 parameters; the 1D-CNN used [32, 64, 128] channels (kernel size 3) and a 50-neuron dense layer, totaling 181,904 parameters.

\subsection{Misclassification Detection and Open Set Recognition}
\label{subsec:uq_eval}

In the CIC-IDS-2017 experiments, DoS attack samples were held out as unknowns and removed from the main dataset. The remaining data was balanced such that benign and known attack samples were equally represented, and split 50/50 into \textit{DEC-Train} and \textit{DEC-Test}. The DEC was trained on \textit{DEC-Train} and used to infer latent features for \textit{DEC-Test} and the held-out DoS set. The \textit{DEC-Test} set was then divided into \textit{XGBoost-Train} and \textit{XGBoost-Test} as an attack recognition task. To train the UQ metamodel, we labeled each sample in \textit{XGBoost-Test} as 0 if correctly predicted and 1 if misclassified, and then subsampled class 0 to five times the size of class 1. This formed a balanced \textit{Metamodel-Train}/\textit{Metamodel-Test} split (80/20). For OSR evaluation, we created the \textit{XGBoost-OSR} set by mixing equal samples from the DoS Holdout and \textit{Metamodel-Test} sets.

In the ACI-IoT-2023 setting, the Slowloris class was designated as the unknown class. The remaining data was balanced and processed in the same pipeline as above: DEC was trained on \textit{DEC-Train}, which excluded Slowloris, and used to transform \textit{DEC-Test} and the Slowloris samples. The resulting latent features were split equally for XGBoost training and testing. Again, we generated the metamodel labels on \textit{XGBoost-Test} and formed the training/test split using a 5:1 correct-to-incorrect ratio. The final OSR test set consisted of an equal mix of Slowloris and \textit{Metamodel-Test} samples.

\section{Results and Discussion}
\label{sec:results}

In this section, we evaluate the effectiveness of the ODXU model and its transferability across NIDS datasets. We employ a two-phase experimental framework: first, ODXU is developed and validated on the CIC-IDS-2017 dataset; then, its generalizability is assessed on the ACI-IoT-2023 dataset using transfer learning techniques. Additionally, we assess UQ methods to detect misclassifications and OSR in both datasets.

\subsection{Results with CIC-IDS-2017 dataset.}
The ODXU model was trained from scratch on the CIC-IDS-2017 dataset \cite{sander2024uncertainty}. The results show that the ODXU model outperforms both FcNN and 1D-CNN on the CIC-IDS-2017 dataset across multiple evaluation metrics. As shown in Table~\ref{tab:obj1results}, ODXU achieves the highest multiclass accuracy at .969 (96.9\%), compared to .934 (93.4\%) and .939 (93.9\%) for FcNN and 1D-CNN, respectively. This improvement is also evident in binary accuracy (.971 for ODXU vs. .946 and .951 for the neural-based models), indicating ODXU's strong ability to distinguish between benign and attacks.

\begin{table}[t!]
\centering
\caption{Results of six metrics across models of the CIC-IDS-2017 dataset.}
\label{tab:obj1results}

\begin{tabular}{|c|c|c|c|}
\hline
\textbf{Measure} & \textbf{FcNN}  & \textbf{1D-CNN} & \textbf{ODXU (Ours)}  \\ \hline
Multiclass Accuracy & .934  & .939 &  \textbf{.969} \\ \hline
Binary Accuracy & .946   & .951 & \textbf{.971} \\ \hline
Misclassified Positive Rate & \textbf{.020}  & .034 & \textbf{.020} \\ \hline
False Omission Rate & .253  & .192 & \textbf{.092} \\ \hline
F1 Score & .967  & .971 & \textbf{.983} \\ \hline
Competence & \textbf{.961}  & .925 & .944 \\ \hline
\end{tabular}
\end{table}

The advantage of ODXU is particularly clear when examining the \textit{False Omission Rate (FOR)}. ODXU achieves a FOR of just .092, a reduction of over 50\% compared to FcNN (.253) and 1D-CNN (.192). This substantial decrease suggests that ODXU is far less likely to overlook actual attacks, making it more reliable for operational cybersecurity deployment.

Additionally, ODXU maintains a strong F1 Score of .983, indicating balanced precision and recall. While the \textit{Misclassified Positive Rate} is tied between ODXU and FcNN (both at .02), the 1D-CNN underperforms slightly at .034, which could lead to more false alarms.

However, the \textit{Competence} score is slightly higher for FcNN (.961) than for ODXU (.944). This supports that while ODXU is more accurate overall, it may be marginally less calibrated regarding predictive certainty. Nonetheless, the trade-off strongly favors ODXU due to its superior accuracy and significantly reduced error in critical classifications.

These results highlight ODXU’s robustness, especially in detecting hard-to-identify attacks such as Bot, DoS Hulk, DOS slowloris, Heartbleed, PortScan, and Web Attacks, which contribute most to its performance gains, refer to Table~\ref{tab:oj1cic2017} for more information.

\subsection{Transferred learning with the ACI-IoT-2023 dataset.}

Table~\ref{tab:vary_dataset} presents the multiclass accuracy of transfer learning models trained on different portions of the ACI-IoT-2023 dataset, based on the configurations defined in Table~\ref{tab:scenarios}. When examining the impact of different DEC configurations, we observe the following performance trend: accuracy improves from models using a fixed pre-trained AE and fine-tuned clustering (e.g., Case~2), to those that fine-tune the AE and train the clustering (e.g., Case~1), and obtains the highest when using a pre-trained AE combined with clustering trained from scratch (e.g., Case~3). For example, with 50\% of the training data, Case~2 achieves an accuracy of .9799, Case~1 improves slightly to .9805, and Case~3 achieves the highest score of .9827.

\begin{table}[t!]
\centering
\caption{Multiclass accuracy of transfer learning across varying training data portions.}
\label{tab:vary_dataset}
\renewcommand{\arraystretch}{1.15}
\begin{tabular}{|c|c|c|c|c|c|c|}
\hline
\textbf{Portion (\%)} & \textbf{Case 1} & \textbf{Case 2} & \textbf{Case 3} & \textbf{Case 4} & \textbf{Case 5} & \textbf{Case 6} \\
\hline
10  & .9666 & .9616 & .9764 & .9680 & .9659 & .9784 \\
\hline
25  & .9690 & .9679 & .9789 & .9736 & .9706 & .9802 \\
\hline
50  & .9805 & .9799 & \textbf{.9827} & \textbf{.9813} & \textbf{.9808} & \textbf{.9841} \\
\hline
75  & \textbf{.9824} & \textbf{.9809} & \textbf{.9836} & \textbf{.9833} & \textbf{.9816} & \textbf{.9845} \\
\hline
\end{tabular}
\end{table}

\begin{figure}[ht!]
    \centering
    \includegraphics[scale=0.32]{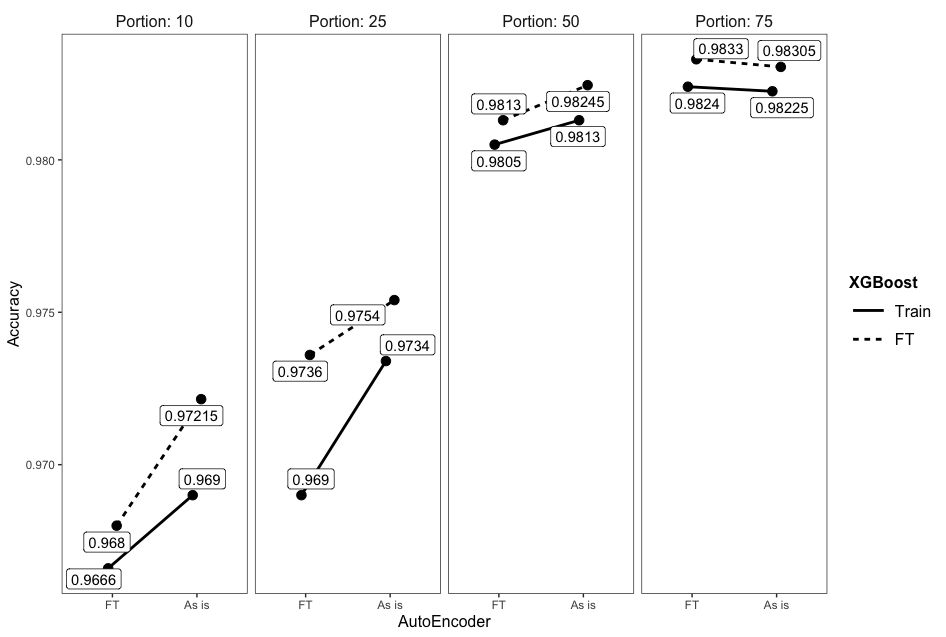}
    \caption{The Accuracy Interaction Plot for different configurations of the ODXU model. For ``FT'' AutoEncoder settings, the plotted accuracy corresponds to the ``Train'' Clustering accuracy from Table~\ref{tab:vary_dataset}. For ``As is'' AutoEncoder settings, the plotted accuracy is the average of the ``FT'' and ``Train'' Clustering accuracies.}
    \label{fig:inter}
\end{figure}

Additionally, when comparing models with identical AE and clustering configurations, we find that fine-tuning the classifier (Cases~4 to 6) consistently obtains higher accuracy than training the classifier from scratch (Cases~1 to 3). For example, with 75\% of the training data, Case~3 achieves .9836 accuracy, whereas Case~6, with a fine-tuned classifier, achieves a higher accuracy of .9845. These results underscore the importance of classifier initialization. Starting from a well-initialized classifier helps to achieve better generalization and performance, emphasizing the advantage of transfer learning. Figure \ref{fig:inter} demonstrates the consistent superior role of fine-tuning the classifier XGBoost in all cases and the success of transfer learning, manifesting in keeping the AE as is.

Transfer learning configurations show superior performance when compared to neural-based models such as FcNN (.9808) and 1D-CNN (.9679). Specifically, Cases~3 to 6 outperform both baselines with as little as 50\% of the training data (approximately 16,000 samples), while Cases~1 and 2 surpass the baselines when trained on 75\% (approximately 23,000 samples)\footnote{These results are based on an ablation study evaluating several portions of data.}. Given that Case~6 (AE: As is, Clustering: Train, Classifier: FT) delivers the best overall accuracy, it is selected for further experimentation in the following sections. 

\paragraph{Early Stopping Optimization}

Initially, all models were trained for the same number of epochs. However, we observed rapid improvements early in training, followed by diminishing returns. To avoid unnecessary computation, we introduced an early stopping mechanism using two hyperparameters: the number of stopping rounds ($\eta$) and thresholds for changes in loss ($\delta$). If the loss during AE pretraining or clustering optimization phases fails to improve by more than $\delta$ over $\eta$ consecutive epochs, training is halted.

We evaluated Case~6 using early stopping rounds $\eta \in [10, 15, 20]$, and two thresholds: $\delta_{\text{AE}} \in [.0005, .001]$ and $\delta_{\text{Cluster}} \in [.005, .01]$. These experiments were conducted with both 50\% and 75\% training data portions.

\begin{table}[ht!]
\centering
\caption{Multiclass accuracy, Precision, Recall, and F1 Score across hyperparameter configurations.}
\label{tab:finetune_config}
\renewcommand{\arraystretch}{1.2}
\begin{tabularx}{\linewidth}{|c|*{6}{>{\centering\arraybackslash}X|}}
\hline
\textbf{Metric} & \textbf{Exp. 1} & \textbf{Exp. 2} & \textbf{Exp. 3} & \textbf{Exp. 4} & \textbf{Exp. 5} & \textbf{Exp. 6} \\
\hline
$\eta$ & 10 & 10 & 15 & 15 & 20 & 20 \\
\hline
$\delta_{\text{AE}}$ & .001 & .0005 & .001 & .0005 & .001 & .0005 \\
\hline
$\delta_{\text{Cluster}}$ & .01 & .005 & .01 & .005 & .01 & .005 \\
\hline
Accuracy (50\%) & .9807 & \textbf{.9817} & \textbf{.9815} & \textbf{.9819} & \textbf{.9820} & \textbf{.9824} \\
\hline
Accuracy (75\%) & .9791 & \textbf{.9820} & .9806 & \textbf{.9831} & \textbf{.9829} & \textbf{.9834} \\
\hline
Precision (50\%) & .9806 & \textbf{.9815} & \textbf{.9813} & \textbf{.9815} & \textbf{.9818} & \textbf{.9823} \\
\hline
Precision (75\%) & .9790 & \textbf{.9818} & \textbf{.9809} & \textbf{.9827} & \textbf{.9828} & \textbf{.9834} \\
\hline
Recall (50\%) & .9807 & \textbf{.9818} & \textbf{.9815} & \textbf{.9819} & \textbf{.9820} & \textbf{.9824} \\
\hline
Recall (75\%) & .9791 & \textbf{.9820} & \textbf{.9811} & \textbf{.9831} & \textbf{.9829} & \textbf{.9834} \\
\hline
F1 Score (50\%) & .9805 & \textbf{.9815} & \textbf{.9813} & \textbf{.9815} & \textbf{.9818} & \textbf{.9823} \\
\hline
F1 Score (75\%) & .9791 & \textbf{.9818} & \textbf{.9809} & \textbf{.9827} & \textbf{.9827} & \textbf{.9834} \\
\hline
Training Time (50\%) & 0:23:07 & 0:25:00 & 0:27:32 & 0:29:10 & 0:39:02 & \textbf{0:49:39} \\
\hline
Training Time (75\%) & 0:20:09 & 0:28:48 & 0:21:38 & 0:37:01 & 0:42:46 & \textbf{1:02:50} \\
\hline
\end{tabularx}
\end{table}

As shown in Table~\ref{tab:finetune_config}, nearly all hyperparameter configurations for both the 50\% and 75\% training portions outperform the FcNN baseline accuracy of .9808, with the exception of Experiment~1. Notably, Experiment~6, featuring the highest early stopping rounds ($\eta = 20$) and the smallest loss thresholds ($\delta_{\text{AE}} = .0005$, $\delta_{\text{Cluster}} = .005$), achieves the best multiclass accuracy at .9824 for 50\% of the training data and .9834 for 75\%. 

This trend of superior performance in Experiment~6 extends consistently across other evaluation metrics, including Precision, Recall, and F1 Score, for both training data proportions. These findings emphasize the value of conservative early stopping parameters, which contribute to stable convergence and enhanced classification performance. In contrast, Experiment~1, configured with the loosest stopping criteria ($\eta = 10$, $\delta_{\text{AE}} = .001$, $\delta_{\text{Cluster}} = .01$), produces the lowest accuracy scores. However, Experiment~1 comes with the benefit of significantly shorter training times: \textbf{0:23:07} for 50\% and \textbf{0:20:09} for 75\% of the data, compared to Experiment~6, which requires \textbf{0:49:39} and \textbf{1:02:50}, respectively (in hh:mm:ss format)\footnote{These training times are wall-clock measurements and may vary depending on the hardware, GPU availability, and runtime environment.}. This contrast highlights the classic trade-off between computational efficiency and model effectiveness.

Based on these results, we selected the configuration from Case~6 with the hyperparameters used in Experiment~6 for further evaluation. The extended performance of this setting on the attack recognition task is presented in Table~\ref{tab:ACI_compare}.

\begin{table}[ht!]
\centering
\caption{Comparison of six evaluation metrics across models.}
\label{tab:ACI_compare}
\renewcommand{\arraystretch}{1.15}
\begin{tabular}{|c|c|c|c|}
\hline
\textbf{Measurement} & \textbf{FcNN} & \textbf{1D-CNN} & \textbf{Case 6, Exp. 6} \\
\hline
Multiclass Accuracy        & .981 & .968 & \textbf{.983} \\
\hline
Binary Accuracy            & .985 & .974 & \textbf{.987} \\
\hline
Misclassified Positive Rate & .022 & .035 & \textbf{.019} \\
\hline
False Omission Rate        & .016 & .029 & \textbf{.014} \\
\hline
F1 Score                   & .985 & .974 & \textbf{.988} \\
\hline
Competence                 & .948 & .935 & \textbf{.969} \\
\hline
\end{tabular}
\end{table}

The table shows that the transfer learning model (Case~6, Exp.~6) achieves the best results across all six evaluation metrics. Similar to the CIC-IDS-2017 dataset, it records the highest multiclass accuracy (.983) and binary accuracy (.987), while also demonstrating the lowest Misclassified Positive Rate (.019) and False Omission Rate (.014). These two metrics are crucial for intrusion detection systems, as they measure how often attacks are misclassified as benign, an outcome with profound security implications. The transfer learning approach also attains the highest F1 score (.988), indicating a strong balance between precision and recall.

Most notably, the \textit{Competence} score, previously slightly higher for FcNN when training a model from the CIC-IDS-2017 dataset, is now also highest for our transfer learning model (.969). This signifies that the model achieves higher accuracy and is better calibrated regarding predictive certainty. In contrast to earlier observations where models like ODXU traded slight drops in competence for gains in accuracy, the current configuration demonstrates that transfer learning can yield improvements in both dimensions.

\begin{figure}[ht!]
    \centering
    \begin{subfigure}{0.49\linewidth}
        \includegraphics[width=\linewidth]{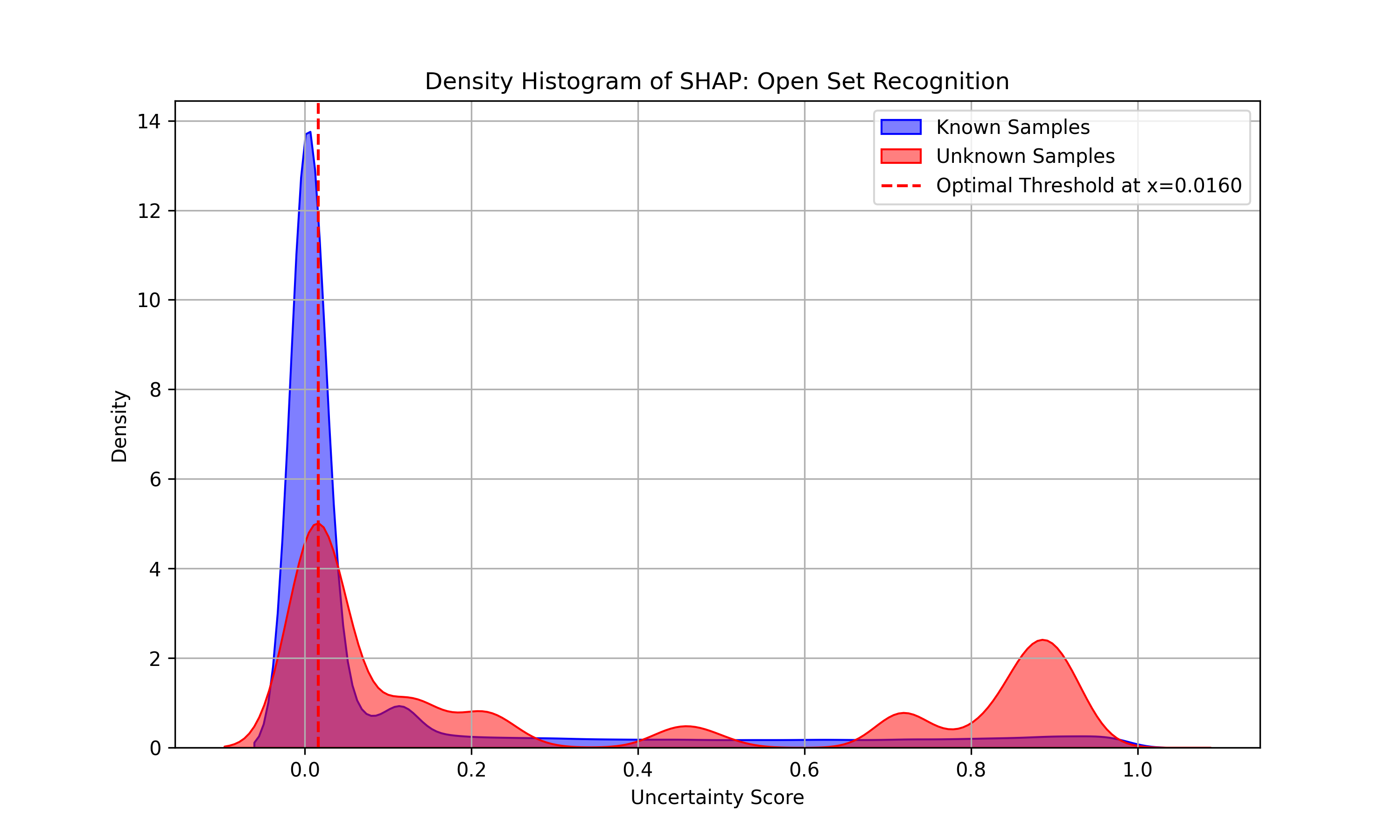}
        \caption{CIC-IDS-2017}
    \end{subfigure}
    \hfill
    \begin{subfigure}{0.49\linewidth}
        \includegraphics[width=\linewidth]{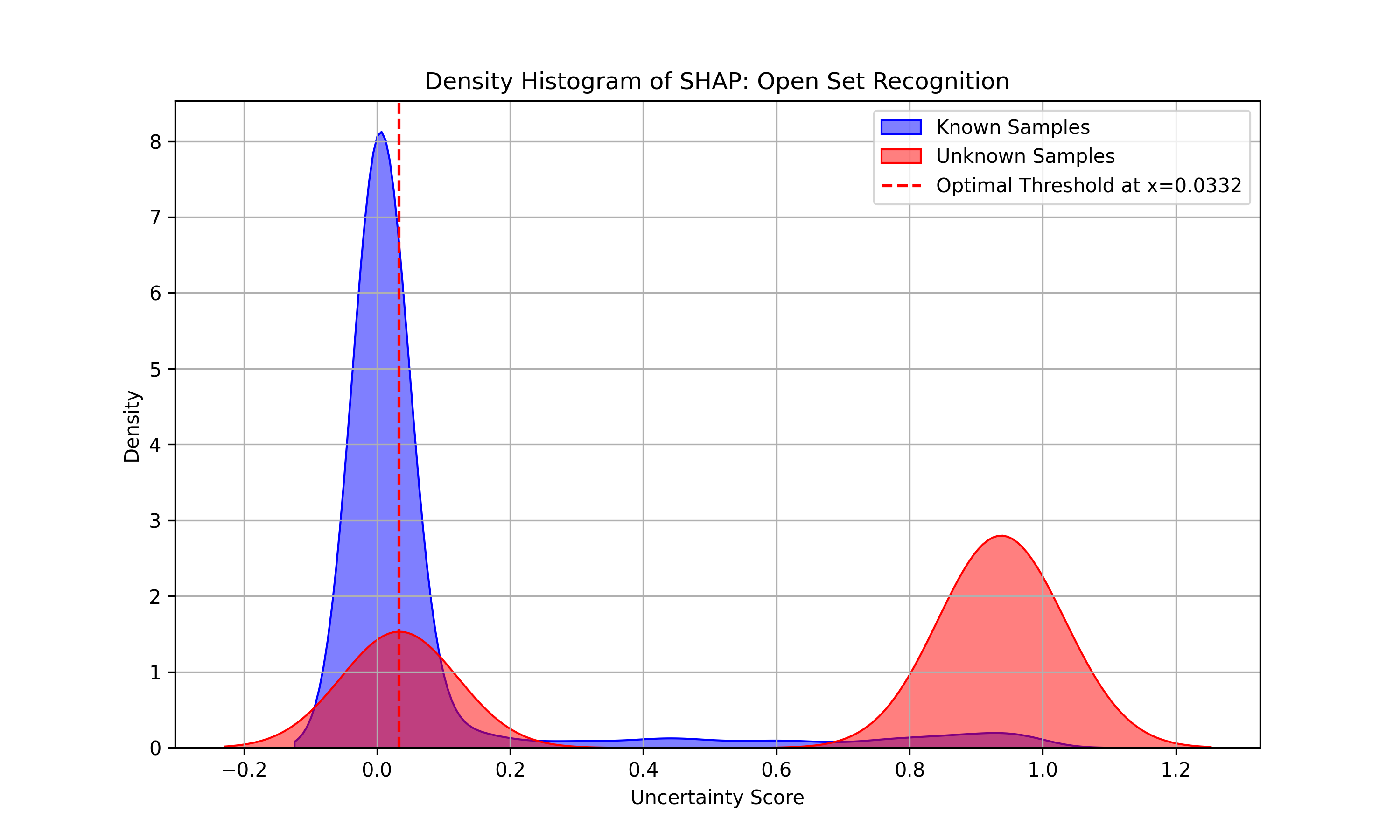}
        \caption{ACI-IoT-2023}
    \end{subfigure}
    \caption{Uncertainty score distributions in the MetaUQ\textsubscript{SHAP} model for known and unknown samples of OSR.}
    \label{fig:uq_density_plot}
\end{figure}

\subsection{Misclassification Detection and Open Set Recognition}

UQ into NIDS offers a powerful way to address challenges like evolving threats, high-dimensional data, and computational efficiency. UQ helps reveal where the NIDS is confident vs. uncertain, which can guide further data collection or model refinement. This section evaluates various UQ methods on two critical tasks: misclassification detection and OSR. 

It is essential to note that UQ models do not directly classify samples as ``known'' or ``unknown.'' Instead, they generate \textit{uncertainty scores} for each sample, which reflect the model’s confidence in its prediction. These scores form a distribution, and by selecting a threshold of this score, we can predict whether the sample belongs to a known or an unknown class. To determine the optimal \textit{uncertainty score} threshold, we use the AUROC, which shows how the true positive rate (TPR) and false positive rate (FPR) change across different threshold values. Samples with uncertainty scores greater than or equal to the chosen threshold are classified as ``unknown.''  Figure~\ref{fig:uq_density_plot} illustrates the distribution of uncertainty scores for both known and unknown samples of $MetaUQ_{\text{SHAP}}$. For the distributions and thresholds of other methods, please refer to the Appendix~\ref{sec:App_B}.

The results, summarized in Tables~\ref{tab:misclassification_compare} and \ref{tab:unknown_compare}, indicate that our proposed metamodel-based approaches consistently outperform score-based methods.
 
\subsubsection{Misclassification Detection}

In the misclassification detection task, all methods perform reasonably well, with AUROC scores approaching or exceeding .9. For the CIC-IDS-2017 dataset, our metamodel variants $\text{MetaUQ}_{\text{Prob}}$, $\text{MetaUQ}_{\text{IG}}$ and $\text{MetaUQ}_{\text{SHAP}}$ achieve AUROCs of .923 and .925, respectively, compared to .895 for both Confidence and Entropy. Similarly, on the ACI-IoT-2023 dataset, all metamodel variants outperform the baselines, with $\text{MetaUQ}_{\text{IG}}$ achieving the highest AUROC of .926.

\begin{table}[ht!]
\centering
\caption{Misclassification performance comparison across UQ methods.}
\label{tab:misclassification_compare}
\renewcommand{\arraystretch}{1.15}
\begin{tabular}{|c|c|c|c|c|c|c|}
\hline
\multirow{2}{*}{\textbf{Dataset}} & \multirow{2}{*}{\textbf{Metric}} & \multicolumn{2}{c|}{\textbf{Score-based}} & \multicolumn{3}{c|}{\textbf{MetaUQ}} \\
\cline{3-7}
& & \textbf{Confidence} & \textbf{Entropy} & \textbf{Prob} & \textbf{SHAP} & \textbf{IG} \\
\hline
\multirow{2}{*}{CIC IDS 2017} 
& AUROC         & .895 & .895 & .923 & \textbf{.925} & .923 \\
& TP@(TN=.95)   & .401 & .400 & .509 & \textbf{.536} & .516 \\
\hline
\multirow{2}{*}{ACI IoT 2023} 
& AUROC         & .911 & .911 & .924 & .924 & \textbf{.926} \\
& TP@(TN=.95)   & .529 & .529 & .559 & .559 & \textbf{.588} \\
\hline
\end{tabular}
\end{table}

The TP@(TN=.95) metric further emphasizes the performance difference. On CIC-IDS-2017, $\text{MetaUQ}_{\text{SHAP}}$ reaches a TP rate of .536, outperforming Confidence (.401) and Entropy (.400). The pattern holds on ACI-IoT-2023, where $\text{MetaUQ}_{\text{IG}}$ achieves a TP rate of .588, a significant improvement over Confidence and Entropy (both at .529). These results suggest that metamodel-based uncertainty estimation is more effective in identifying misclassified predictions, making them highly useful for real-world deployment.

\subsubsection{Open Set Recognition (OSR)}

On CIC-IDS-2017, $\text{MetaUQ}_{\text{SHAP}}$ attains the highest AUROC of .745, surpassing both score-based methods and other metamodels. On ACI-IoT-2023, $\text{MetaUQ}_{\text{SHAP}}$ also leads with an AUROC of .938, significantly ahead of the next best, $\text{MetaUQ}_{\text{Prob}}$ and $\text{MetaUQ}_{\text{IG}}$ (both at .921).

\begin{table}[ht!]
\centering
\caption{OSR Unknown attack performance comparison across UQ methods.}
\label{tab:unknown_compare}
\renewcommand{\arraystretch}{1.15}
\begin{tabular}{|c|c|c|c|c|c|c|}
\hline
\multirow{2}{*}{\textbf{Dataset}} & \multirow{2}{*}{\textbf{Metric}} & \multicolumn{2}{c|}{\textbf{Score-based}} & \multicolumn{3}{c|}{\textbf{MetaUQ}} \\
\cline{3-7}
& & \textbf{Confidence} & \textbf{Entropy} & \textbf{Prob} & \textbf{SHAP} & \textbf{IG} \\
\hline
\multirow{2}{*}{CIC IDS 2017} 
& AUROC         & .710 & .723 & .728 & \textbf{.745} & .725 \\
& TP@(TN=.95)   & \textbf{.212} & \textbf{.212} & .145 & .145 & .142 \\
\hline
\multirow{2}{*}{ACI IoT 2023} 
& AUROC         & .916 & .919 & .921 & \textbf{.938} & .921 \\
& TP@(TN=.95)   & .435 & .462 & .489 & \textbf{.590} & .469 \\
\hline
\end{tabular}
\end{table}

In terms of TP@(TN=.95), the advantage of $\text{MetaUQ}_{\text{SHAP}}$ is especially pronounced. On ACI-IoT-2023, it achieves a TP rate of .590, exceeding the second-best method ($\text{MetaUQ}_{\text{Prob}}$ at .489) by more than 10\%. This large margin highlights the capability of SHAP-based explanations to uncover high-quality uncertainty signals that generalize to unseen attack types.

\section{Conclusion}
\label{sec:conclusion}

This paper presents an extension of the Neurosymbolic AI framework applied in network intrusion detection systems, named Open Set Recognition with Deep Embedded Clustering for XGBoost and Uncertainty Quantification (ODXU). First, we present the results on the CIC-IDS-2017 dataset, demonstrating that the hybrid architecture outperforms neural-based models such as FcNN and 1D-CNN across most evaluation metrics. This suggests the effectiveness of combining neural feature extraction with symbolic reasoning for robust and interpretable intrusion detection.

Second, we propose a transfer learning paradigm of ODXU and evaluate its generalizability on the ACI-IoT-2023 dataset. Our ablation studies show that a model trained on a large, well-structured dataset (CIC-IDS-2017) can be successfully adapted to a new target dataset (ACI-IoT-2023), achieving high performance with fewer training data. The optimal transfer configuration involves reusing the pre-trained Autoencoder (AE), training the clustering module from scratch, and fine-tuning the XGBoost classifier. This setup outperforms neural-based models when trained on around 50\% of the target data (approximately 16,000 samples), balancing model performance and data efficiency.

To reduce computational overhead and prevent overfitting, we implement an early stopping strategy, halting training if the Autoencoder and clustering losses fall below 0.0005 and 0.005, respectively, for 20 consecutive epochs. This optimization maintains high accuracy while improving training efficiency.

We also evaluate five Uncertainty Quantification (UQ) methods, two scoring-based and three metamodel-based, on both misclassification detection and Open Set Recognition (OSR) tasks for both datasets. Our results show that metamodel-based approaches consistently outperform scoring-based ones in both tasks. However, the optimal metamodel varies depending on the dataset and evaluation objective. For misclassification detection, $\text{MetaUQ}_{\text{SHAP}}$ achieves the best performance on CIC-IDS-2017, while $\text{MetaUQ}_{\text{IG}}$ performs best on ACI-IoT-2023. In contrast, for OSR, $\text{MetaUQ}_{\text{SHAP}}$ outperforms all other methods across both datasets. These findings highlight the importance of selecting UQ methods that are well-suited to the specific detection task and data distribution.

In future work, we plan to extend our framework to additional datasets such as the Canadian Institute for Cybersecurity IoT 2023 dataset (CIC-IoT-2023) and the Unified Multimodal Network Intrusion Detection Systems (UM-NIDS) dataset. We will also explore out-of-distribution detection tasks to further validate the robustness and generalizability of the proposed transfer learning approach.

\section*{Acknowledgment}
This work was supported by the U.S. Military Academy (USMA) under Cooperative Agreement No. W911NF-23-2-0108 and the Defense Advanced Research Projects Agency (DARPA) under Support Agreement No. USMA 23004. The views and conclusions expressed in this paper are those of the authors and do not reflect the official policy or position of the U.S. Military Academy, U.S. Army, U.S. Department of Defense, or U.S. Government.
	
\bibliographystyle{unsrt}   
\bibliography{main}

\begin{thebibliography}{10}

\bibitem{al2021intelligent}
Mohammad Al-Omari, Majdi Rawashdeh, Fadi Qutaishat, Mohammad Alshira’H, and Nedal Ababneh.
\newblock An intelligent tree-based intrusion detection model for cyber security.
\newblock {\em Journal of Network and Systems Management}, 29(2):20, 2021.

\bibitem{sander2024uncertainty}
Jacob Sander, Chung-En~Johnny Yu, Brian Jalaian, and Nathaniel~D. Bastian.
\newblock Uncertainty-quantified neurosymbolic ai for open set recognition in network intrusion detection.
\newblock In {\em 2024 IEEE Military Communications Conference (MILCOM)}, pages 13--18, Washington, DC, USA, 2024. IEEE.

\bibitem{jalaian2023neurosymbolic}
Brian Jalaian and Nathaniel~D. Bastian.
\newblock Neurosymbolic ai in cybersecurity: Bridging pattern recognition and symbolic reasoning.
\newblock In {\em Proceedings of the 2023 IEEE Military Communications Conference (MILCOM)}, pages 268--273, Boston, MA, USA, 2023. IEEE.

\bibitem{shahoveisi2023application}
Fereshteh Shahoveisi, Hamed Taheri~Gorji, Seyedmojtaba Shahabi, Seyedali Hosseinirad, Samuel Markell, and Fartash Vasefi.
\newblock Application of image processing and transfer learning for the detection of rust disease.
\newblock {\em Scientific Reports}, 13(1):5133, 2023.

\bibitem{amiriparian2022deepspectrumlite}
Shahin Amiriparian, Tobias H{\"u}bner, Vincent Karas, Maurice Gerczuk, Sandra Ottl, and Bj{\"o}rn~W Schuller.
\newblock Deepspectrumlite: A power-efficient transfer learning framework for embedded speech and audio processing from decentralized data.
\newblock {\em Frontiers in Artificial Intelligence}, 5:856232, 2022.

\bibitem{moon2014multimodal}
Seungwhan Moon, Suyoun Kim, and Haohan Wang.
\newblock Multimodal transfer deep learning with applications in audio-visual recognition.
\newblock \url{https://arxiv.org/abs/1412.3121}, 2014.
\newblock arXiv:1412.3121 [cs.LG].

\bibitem{kim2022transfer}
Hee~E Kim, Alejandro Cosa-Linan, Nandhini Santhanam, Mahboubeh Jannesari, Mate~E Maros, and Thomas Ganslandt.
\newblock Transfer learning for medical image classification: a literature review.
\newblock {\em BMC medical imaging}, 22(1):69, 2022.

\bibitem{denning1987intrusion}
Dorothy~E Denning.
\newblock An intrusion-detection model.
\newblock {\em IEEE Transactions on software engineering}, SE-13(2):222--232, 1987.

\bibitem{Yang2024}
Jingkang Yang, Kaiyang Zhou, Yixuan Li, and Ziwei Liu.
\newblock Generalized out-of-distribution detection: A survey.
\newblock {\em International Journal of Computer Vision}, 132(12):5635–5662, Jun 2024.

\bibitem{farrukh2022payloadbyte}
Yasir~Ali Farrukh, Irfan Khan, Syed Wali, David Bierbrauer, John~A. Pavlik, and Nathaniel~D. Bastian.
\newblock Payload-byte: A tool for extracting and labeling packet capture files of modern network intrusion detection datasets.
\newblock In {\em Proceedings of the 2022 IEEE/ACM International Conference on Big Data Computing, Applications and Technologies (BDCAT)}, pages 58--67, Vancouver, BC, Canada, 2022. IEEE.

\bibitem{moustafa2015unsw}
Nour Moustafa and Jill Slay.
\newblock Unsw-nb15: A comprehensive data set for network intrusion detection systems (unsw-nb15 network data set).
\newblock In {\em Proceedings of the 2015 Military Communications and Information Systems Conference (MilCIS)}, pages 1--6, Canberra, ACT, Australia, 2015. IEEE.

\bibitem{sharafaldincicids2017}
Iman Sharafaldin, Arash~Habibi Lashkari, Ali~A Ghorbani, et~al.
\newblock Toward generating a new intrusion detection dataset and intrusion traffic characterization.
\newblock {\em ICISSp}, 1:108--116, 2018.

\bibitem{Nack_2024b}
Emily Nack.
\newblock Aci iot network traffic dataset 2023, Apr 2024.

\bibitem{bizzarri2024synergistic}
Alice Bizzarri, Chung-En Yu, Brian Jalaian, Fabrizio Riguzzi, and Nathaniel~D. Bastian.
\newblock A synergistic approach in network intrusion detection by neurosymbolic ai.
\newblock \url{https://arxiv.org/abs/2406.00938}, 2024.
\newblock arXiv:2406.00938 [cs.CR].

\bibitem{lee2022surgical}
Yoonho Lee, Annie~S. Chen, Fahim Tajwar, Ananya Kumar, Huaxiu Yao, Percy Liang, and Chelsea Finn.
\newblock Surgical fine-tuning improves adaptation to distribution shifts.
\newblock \url{https://arxiv.org/abs/2210.11466}, 2022.
\newblock arXiv:2210.11466 [cs.LG].

\bibitem{liu2021transtailor}
Bingyan Liu, Yifeng Cai, Yao Guo, and Xiangqun Chen.
\newblock Transtailor: Pruning the pre-trained model for improved transfer learning.
\newblock In {\em Proceedings of the AAAI Conference on Artificial Intelligence}, volume~35, pages 8627--8634, Vancouver, BC, Canada (held virtually), 2021. AAAI Press.

\bibitem{guo2019spottune}
Yunhui Guo, Honghui Shi, Abhishek Kumar, Kristen Grauman, Tajana Rosing, and Rogerio Feris.
\newblock Spottune: Transfer learning through adaptive fine-tuning.
\newblock In {\em Proceedings of the IEEE/CVF Conference on Computer Vision and Pattern Recognition (CVPR)}, pages 4805--4814, Long Beach, CA, USA, 2019. IEEE.

\bibitem{vadera2020ursabench}
Meet~P. Vadera, Adam~D. Cobb, Brian Jalaian, and Benjamin~M. Marlin.
\newblock Ursabench: Comprehensive benchmarking of approximate bayesian inference methods for deep neural networks.
\newblock \url{https://arxiv.org/abs/2007.04466}, 2020.
\newblock arXiv:2007.04466 [cs.LG].

\bibitem{hendrycks2016misclassificationood}
Dan Hendrycks and Kevin Gimpel.
\newblock A baseline for detecting misclassified and out-of-distribution examples in neural networks.
\newblock \url{https://arxiv.org/abs/1610.02136}, 2016.
\newblock arXiv:1610.02136 [cs.LG].

\bibitem{wong2023uncertaintywintersim}
Joshua~A. Wong, Alexander~M. Berenbeim, David~A. Bierbrauer, and Nathaniel~D. Bastian.
\newblock Uncertainty-quantified, robust deep learning for network intrusion detection.
\newblock In {\em Proceedings of the 2023 Winter Simulation Conference (WSC)}, pages 2470--2481, San Antonio, TX, USA, 2023. IEEE.

\bibitem{ghosh2021ibmuncertainty}
Soumya Ghosh, Q.~Vera Liao, Karthikeyan~Natesan Ramamurthy, Jiri Navratil, Prasanna Sattigeri, Kush~R. Varshney, and Yunfeng Zhang.
\newblock Uncertainty quantification 360: A holistic toolkit for quantifying and communicating the uncertainty of ai.
\newblock \url{https://arxiv.org/abs/2106.01410}, 2021.
\newblock arXiv:2106.01410 [cs.LG].

\bibitem{chen2019ibmwhitebox}
Tongfei Chen, Jiří Navrátil, Vijay Iyengar, and Karthikeyan Shanmugam.
\newblock Confidence scoring using whitebox meta-models with linear classifier probes.
\newblock In {\em Proceedings of the 22nd International Conference on Artificial Intelligence and Statistics (AISTATS 2019)}, pages 1467--1475, Naha, Okinawa, Japan, 2019. Proceedings of Machine Learning Research.

\bibitem{lundberg2017unified}
Scott~M. Lundberg and Su-In Lee.
\newblock A unified approach to interpreting model predictions.
\newblock {\em Advances in Neural Information Processing Systems}, 30:4765--4774, 2017.

\bibitem{quinlan1986induction}
J.~Ross Quinlan.
\newblock Induction of decision trees.
\newblock {\em Machine learning}, 1:81--106, 1986.

\bibitem{xie2016dec}
Junyuan Xie, Ross Girshick, and Ali Farhadi.
\newblock Unsupervised deep embedding for clustering analysis.
\newblock In {\em Proceedings of the 33rd International Conference on Machine Learning (ICML)}, ICML'16, pages 478--487, New York, NY, USA, 2016. JMLR.org.

\bibitem{chen2016xgboost}
Tianqi Chen and Carlos Guestrin.
\newblock Xgboost: A scalable tree boosting system.
\newblock In {\em Proceedings of the 22nd ACM SIGKDD International Conference on Knowledge Discovery and Data Mining (KDD '16)}, pages 785--794, San Francisco, CA, USA, 2016. ACM.

\bibitem{smith2011quantifying}
Geoffrey Smith.
\newblock Quantifying information flow using min-entropy.
\newblock In {\em Proceedings of the 2011 Eighth International Conference on Quantitative Evaluation of Systems (QEST)}, pages 159--167, Aachen, Germany, 2011. IEEE.

\bibitem{tagasovska2019single}
Natasa Tagasovska and David Lopez-Paz.
\newblock Single-model uncertainties for deep learning.
\newblock {\em Advances in Neural Information Processing Systems}, 32:6415--6425, 2019.

\bibitem{lundberg_shap}
Scott~M. Lundberg, Gabriel~G. Erion, and Su-In Lee.
\newblock Consistent individualized feature attribution for tree ensembles, 2019.

\bibitem{lundberg2020local}
Scott~M Lundberg, Gabriel Erion, Hugh Chen, Alex DeGrave, Jordan~M Prutkin, Bala Nair, Ronit Katz, Jonathan Himmelfarb, Nisha Bansal, and Su-In Lee.
\newblock From local explanations to global understanding with explainable ai for trees.
\newblock {\em Nature machine intelligence}, 2(1):56--67, 2020.

\bibitem{bradley1997auroc}
Andrew~P Bradley.
\newblock The use of the area under the roc curve in the evaluation of machine learning algorithms.
\newblock {\em Pattern recognition}, 30(7):1145--1159, 1997.

\bibitem{matejek2024safeguarding}
Brian Matejek, Ashish Gehani, Nathaniel~D. Bastian, Daniel Clouse, Bradford Kline, and Susmit Jha.
\newblock Safeguarding network intrusion detection models from zero-day attacks and concept drift.
\newblock In {\em Proceedings of the AAAI Workshop on Artificial Intelligence for Cyber Security (AICS)}, pages 1--12, Vancouver, BC, Canada, 2024. AAAI Press.

\bibitem{aciiot2023}
Nathaniel Bastian, David Bierbrauer, Morgan McKenzie, and Emily Nack.
\newblock {ACI IoT Network Traffic Dataset 2023}.
\newblock \url{https://dx.doi.org/10.21227/qacj-3x32}, 2023.
\newblock DOI: 10.21227/qacj-3x32.

\end{thebibliography}

\appendix
\section{Per-class accuracy}

\begin{table}[H]
\centering
\caption{Per-class accuracy across models of the CIC-IDS-2017.}
\label{tab:oj1cic2017}
\begin{tabular}{|c|c|c|c|}
\hline
\textbf{Class Accuracy} & \textbf{FcNN} & \textbf{1D-CNN} & \textbf{ODXU (Ours)}\\ \hline
All-Class & .934 & .939 & \textbf{.969} \\ \hline
Benign & \textbf{.974} & .889 & .902 \\ \hline
BOT & .642 & .717 & \textbf{.909} \\ \hline
DDoS & \textbf{1} & \textbf{1} & \textbf{1} \\ \hline
DoS Goldeneye & \textbf{.998} & .983 & .985 \\ \hline
DoS Hulk & .894 & .894 & \textbf{.998} \\ \hline
DoS Slowhttptest & \textbf{.997} & .996 & .993\\ \hline
DoS slowloris &.996 & .997 & \textbf{.998} \\ \hline
FTP-Patator  & .999 & .999 & .999 \\ \hline
Heartbleed & .141 & .471 & \textbf{.770} \\ \hline
Infiltration & \textbf{.999} & \textbf{.999} & .997 \\ \hline
Portscan & .411 & .411 & \textbf{.548} \\ \hline
SSH-Patator & \textbf{.999}  & \textbf{.999} & .996 \\ \hline
Web Attack Brute Force & .998  & .998 & \textbf{.999} \\ \hline
Web Attack SQL Injection & \textbf{.976}  & .952 & .868 \\ \hline
Web Attack XSS & .987  & .989 & \textbf{.992} \\ \hline

\end{tabular}
\end{table}

\begin{table}[H]
\centering
\caption{Per-class accuracy across models on the ACI-IoT-2023 dataset.}
\label{tab:obj1aci2023}
\begin{tabular}{|c|c|c|c|}
\hline
\textbf{Class} & \textbf{FcNN} & \textbf{1D-CNN} & \textbf{Case~6, Exp.~6} \\
\hline
Benign & .986 & .976 & \textbf{.989} \\
\hline
Port Scan & .656 & .669 & \textbf{.738} \\
\hline
Dictionary Attack & .994 & \textbf{.995} & .991 \\
\hline
SYN Flood & \textbf{1} & \textbf{1} & \textbf{1} \\
\hline
DNS Flood & \textbf{.994} & .990 & .993 \\
\hline
Slowloris & \textbf{1} & \textbf{1} & \textbf{1} \\
\hline
OS Scan & .764 & .566 & \textbf{.808} \\
\hline
ICMP Flood & \textbf{1} & \textbf{1} & \textbf{1} \\
\hline
UDP Flood & \textbf{.842} & .632 & .841 \\
\hline
Vulnerability Scan & \textbf{.816} & .610 & .725 \\
\hline
\end{tabular}
\end{table}
\newpage

\section{The distributions and thresholds across UQ methods for OSR task}
\label{sec:App_B}
\vspace{-10pt}
\begin{figure}[ht!]
\centering
\begin{subfigure}{0.47\linewidth}
  \includegraphics[width=\linewidth]{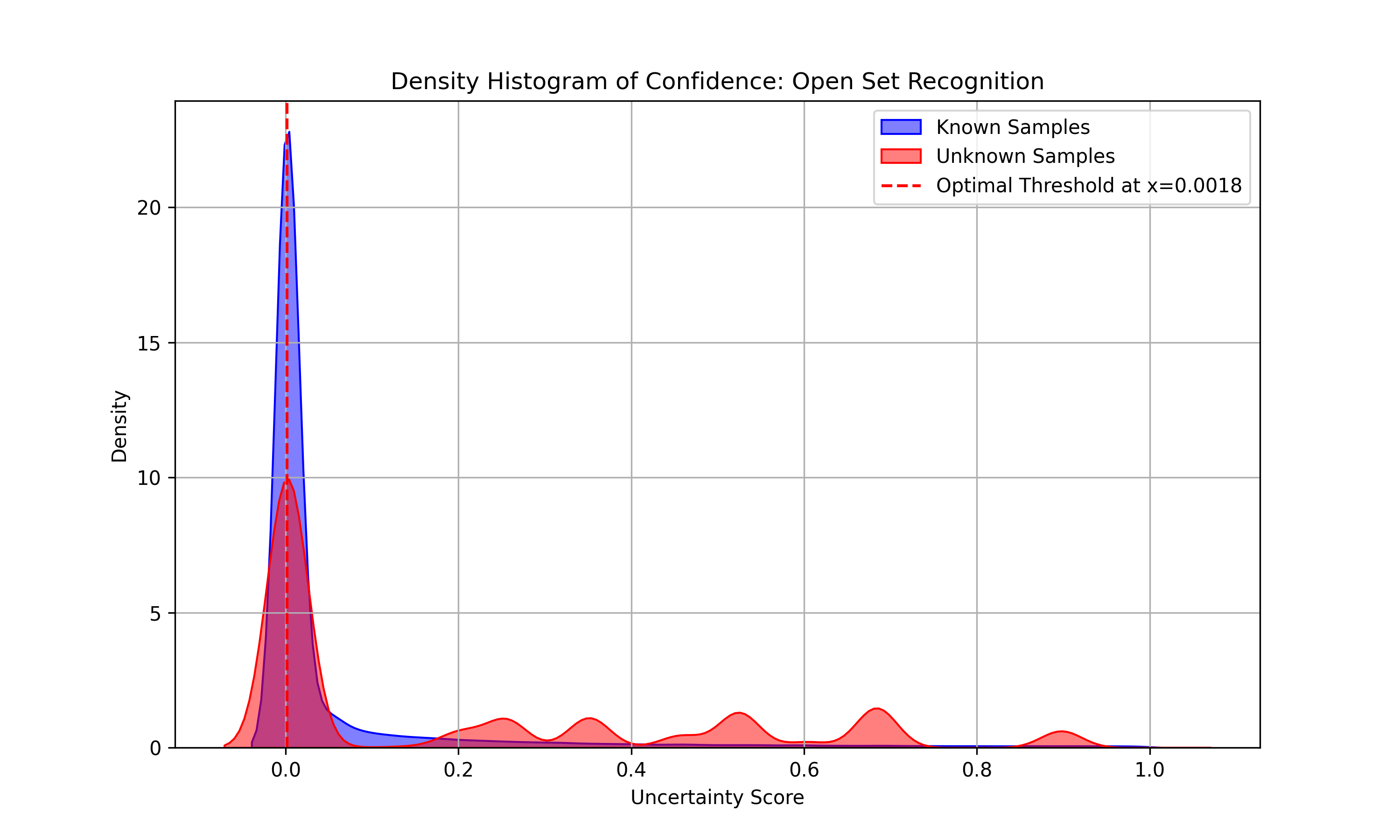}
  \caption{Confidence Scoring (CIC-IDS-2017)}
\end{subfigure}
\hfill
\begin{subfigure}{0.47\linewidth}
  \includegraphics[width=\linewidth]{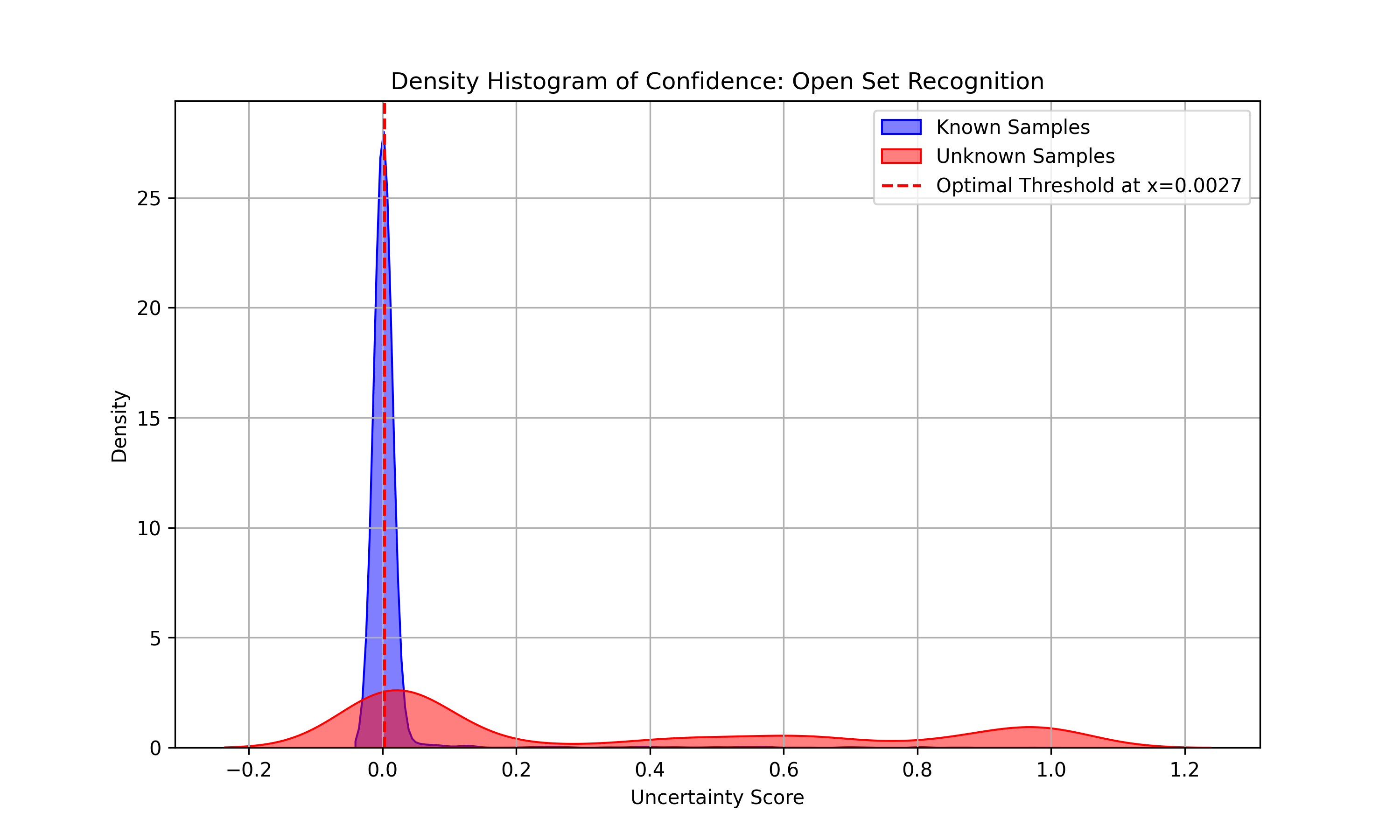}
  \caption{Confidence Scoring (ACI-IoT-2023)}
\end{subfigure}

\begin{subfigure}{0.47\linewidth}
  \includegraphics[width=\linewidth]{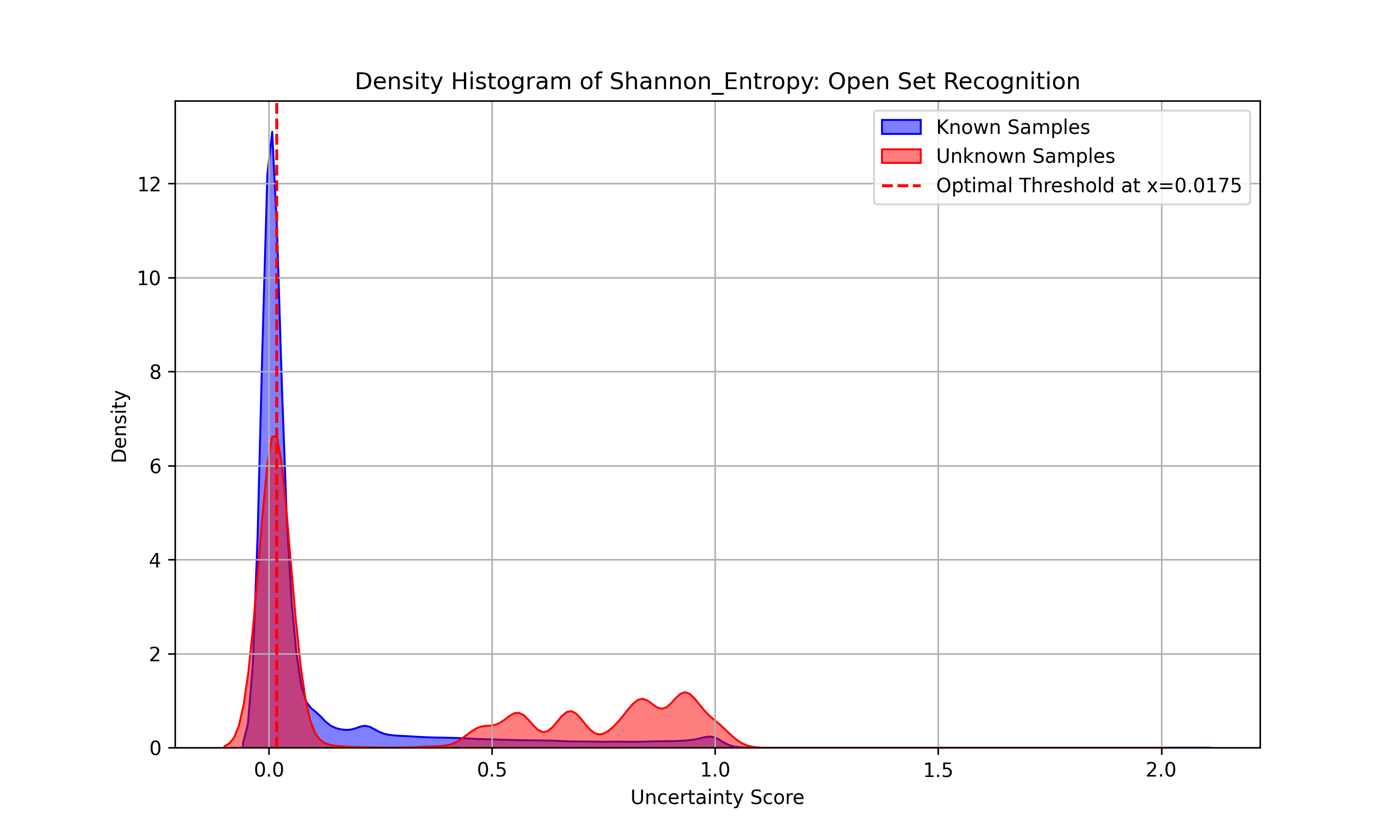}
  \caption{Shannon Entropy (CIC-IDS-2017)}
\end{subfigure}
\hfill
\begin{subfigure}{0.47\linewidth}
  \includegraphics[width=\linewidth]{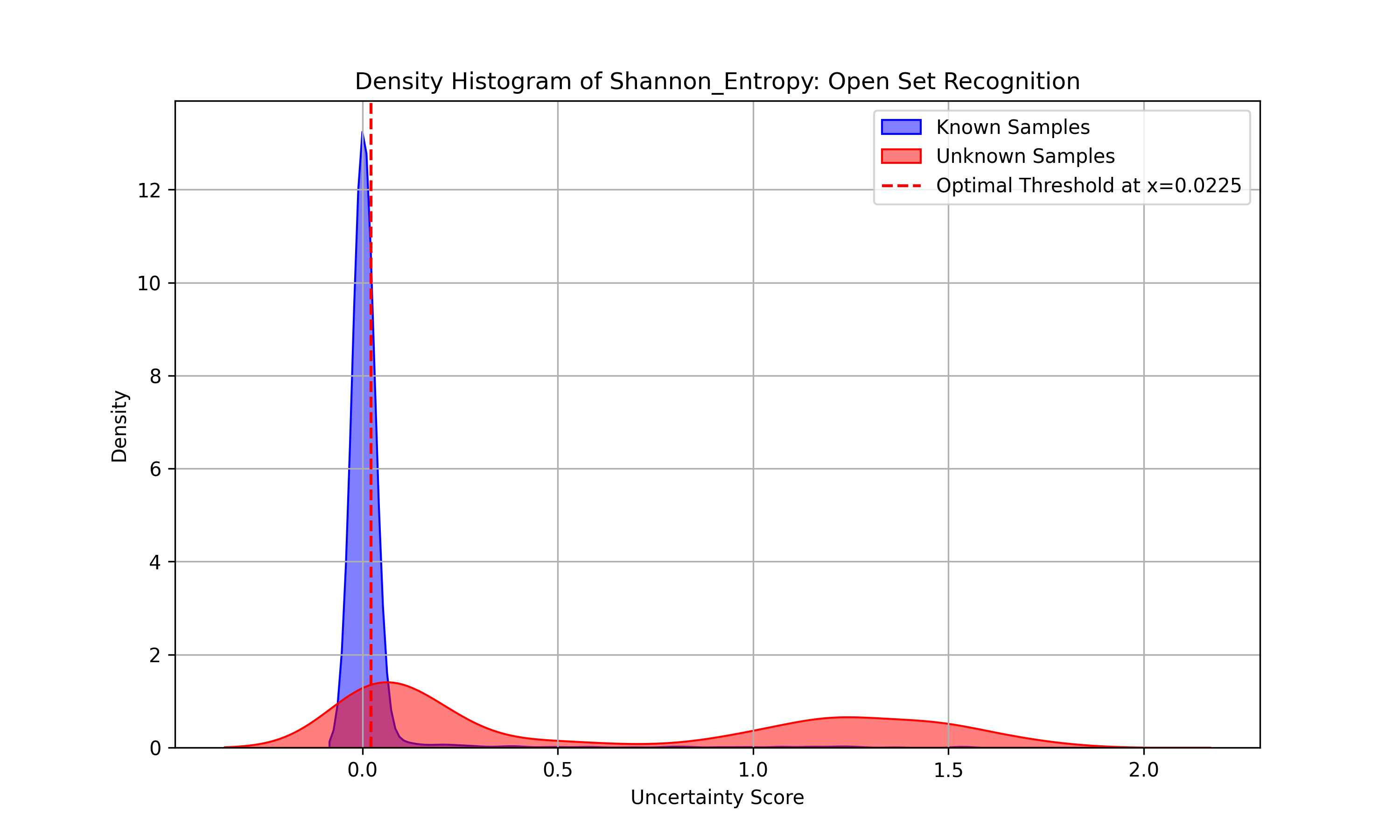}
  \caption{Shannon Entropy (ACI-IoT-2023)}
\end{subfigure}

\begin{subfigure}{0.47\linewidth}
  \includegraphics[width=\linewidth]{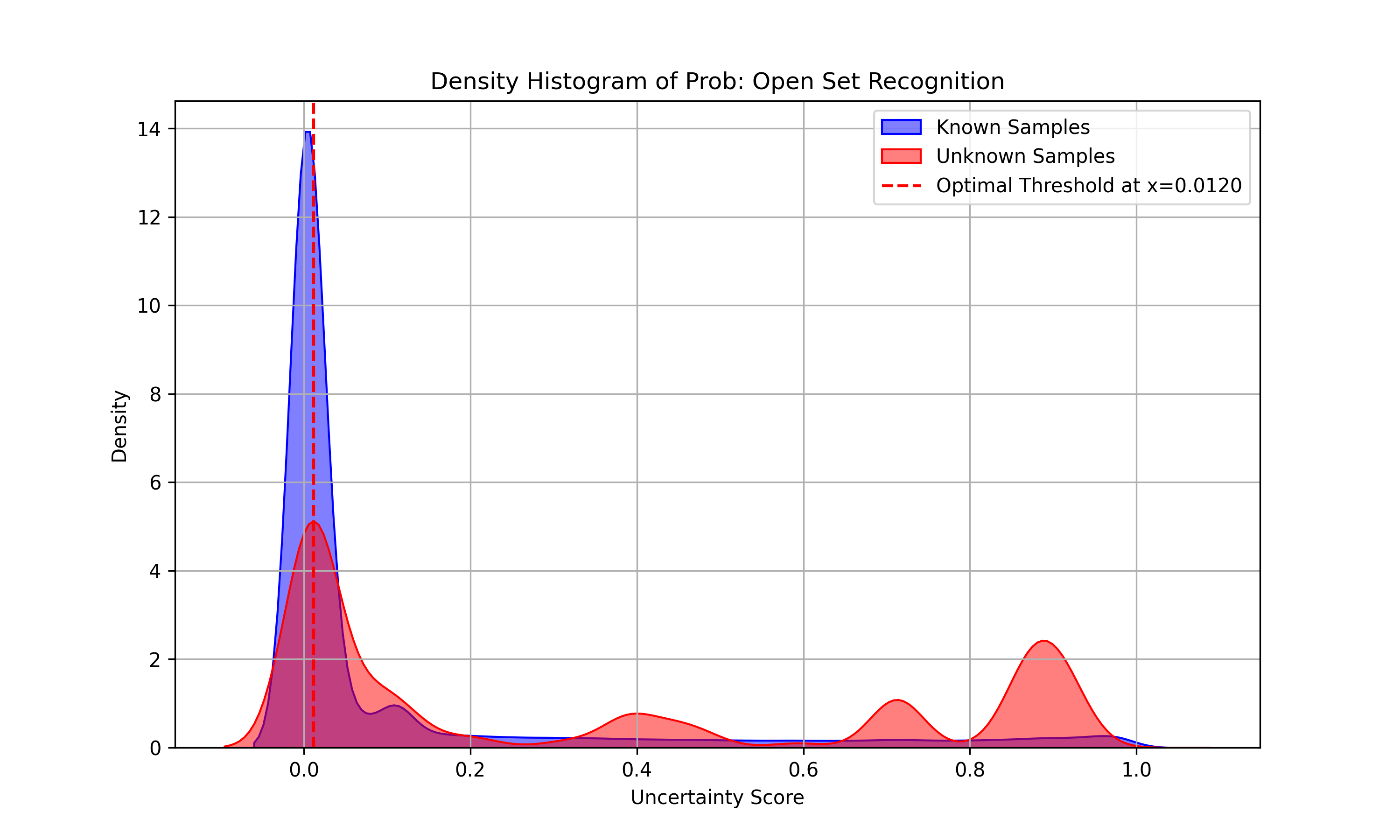}
  \caption{$\text{MetaUQ}_{\text{Prob}}$ (CIC-IDS-2017)}
\end{subfigure}
\hfill
\begin{subfigure}{0.47\linewidth}
  \includegraphics[width=\linewidth]{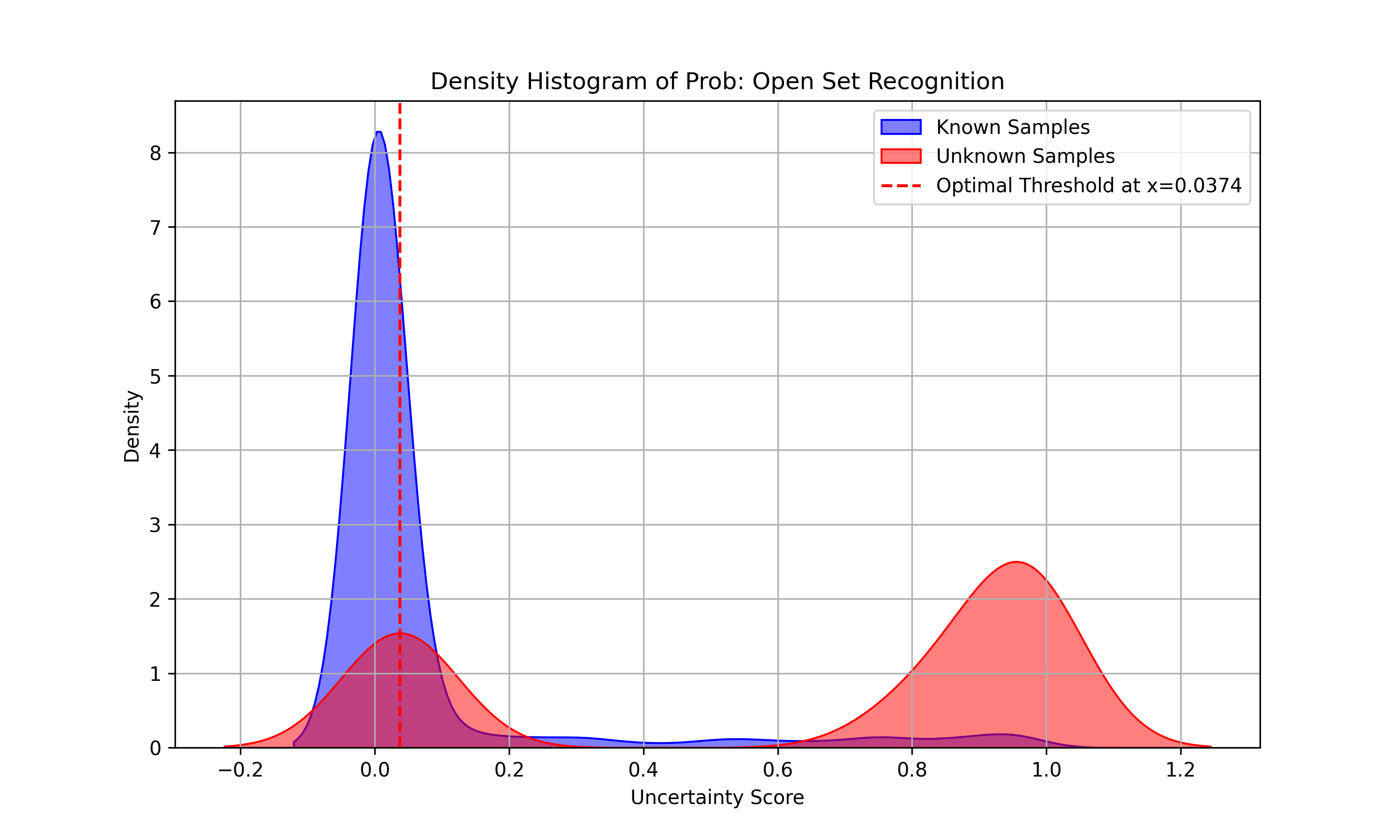}
  \caption{$\text{MetaUQ}_{\text{Prob}}$ (ACI-IoT-2023)}
\end{subfigure}

\begin{subfigure}{0.47\linewidth}
  \includegraphics[width=\linewidth]{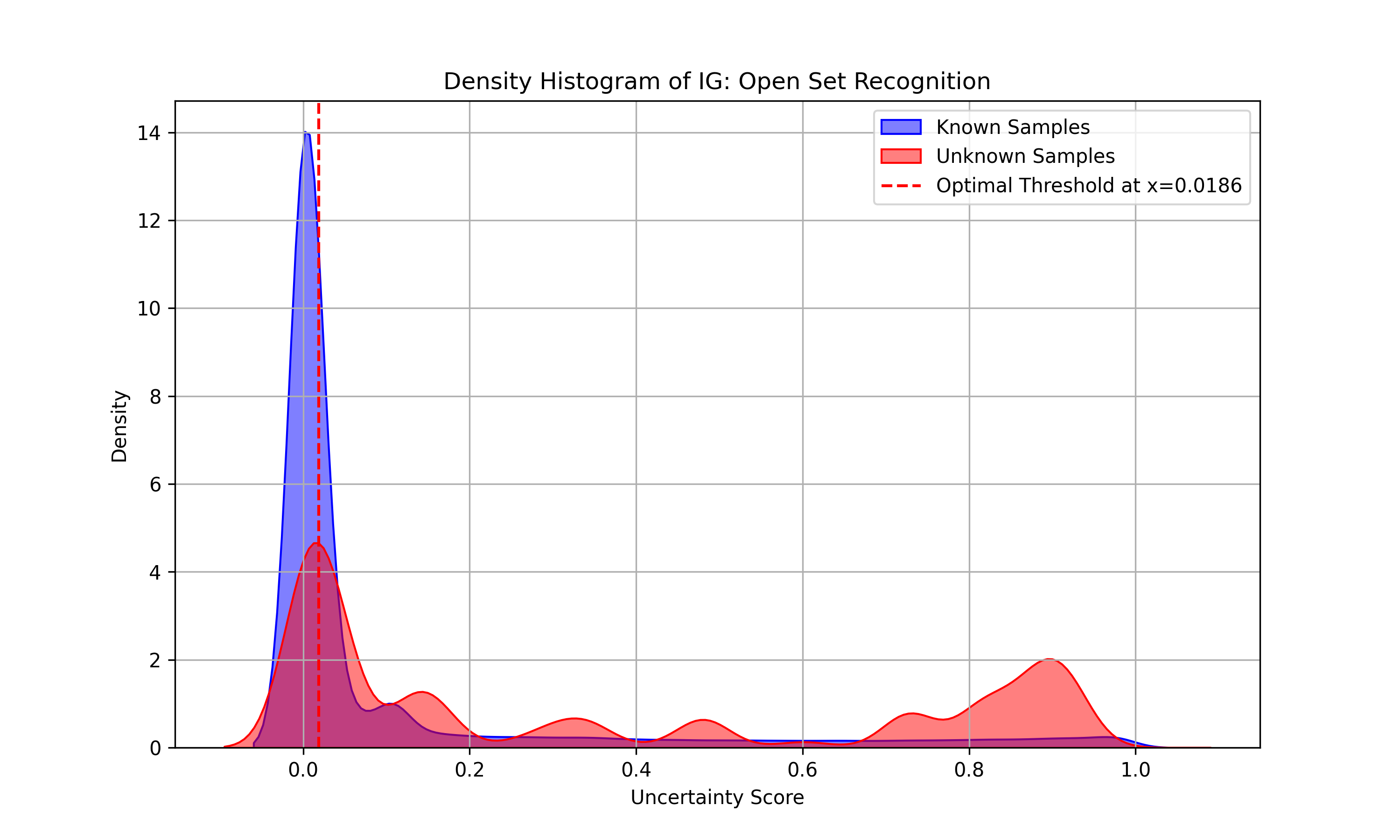}
  \caption{$\text{MetaUQ}_{\text{IG}}$ (CIC-IDS-2017)}
\end{subfigure}
\hfill
\begin{subfigure}{0.47\linewidth}
  \includegraphics[width=\linewidth]{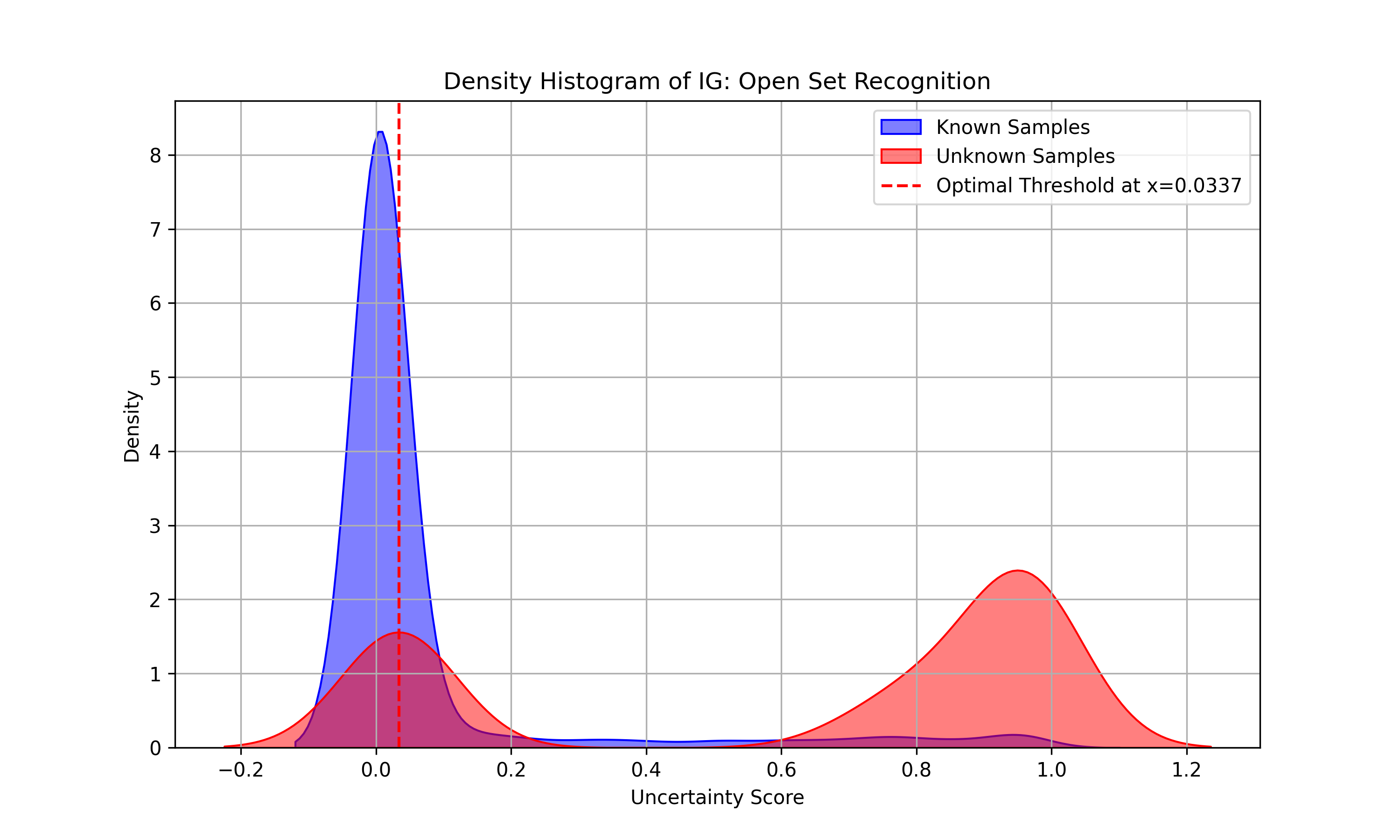}
  \caption{$\text{MetaUQ}_{\text{IG}}$ (ACI-IoT-2023)}
\end{subfigure}
\vspace{-5pt}
\caption{Comparison of uncertainty score distributions. Column 1: CIC-IDS-2017. Column 2: ACI-IoT-2023.
Row 1: Confidence scoring, Row 2: Shannon Entropy, Row 3: $\text{MetaUQ}_{\text{Prob}}$, Row 4: $\text{MetaUQ}_{\text{IG}}$.}
\label{fig:uq_method_comparison}
\end{figure}

\end{document}